\documentclass[a4paper,american,reqno]{amsart}

\usepackage{babel}
\usepackage[utf8]{inputenc}
\usepackage[T1]{fontenc}
\usepackage{xspace}
\usepackage{todonotes}
\usepackage{booktabs}
\usepackage{adjustbox}
\usepackage{multirow}
\usepackage[style = authoryear-comp,
            maxbibnames = 100,
            maxcitenames = 2,
            giveninits = true,
            uniquename = init,
            isbn = false,
            backend = bibtex]{biblatex}
\usepackage[colorlinks,
            citecolor=blue,
            urlcolor=blue,
            linkcolor=blue]{hyperref} 

\makeatletter
\patchcmd{\@settitle}{\uppercasenonmath\@title}{\scshape\large}{}{}
\patchcmd{\@setauthors}{\MakeUppercase}{\scshape\normalsize}{}{}
\makeatother

\renewenvironment{description}
  {\list{}{%
    \setlength{\leftmargin}{1.5em}%
    \setlength{\itemindent}{-\leftmargin}%
    \setlength{\labelwidth}{0pt}%
    }}
  {\endlist}

\vfuzz \hfuzz
\makeatletter
\@namedef{subjclassname@2020}{%
  \textup{2020} Mathematics Subject Classification}
\makeatother

\newtheorem{prop}{Proposition}

\newtheorem{thm}{Theorem}
\newtheorem{lem}[thm]{Lemma}

\newcommand{\nint}[1]{\left\lfloor #1 \right\rceil}

\newcommand{\st}{\text{s.t.}}
\newcommand{\define}{\mathrel{{\mathop:}{=}}}

\makeatletter
\newenvironment{varsubequations}[1]
{%
  \addtocounter{equation}{-1}%
  \renewcommand{\theHequation}{formulation.#1}%
  \begin{subequations}
    \renewcommand{\theparentequation}{#1}%
    \def\@currentlabel{#1}%

  }
  {%
  \end{subequations}\ignorespacesafterend
}
\makeatother

\newcommand\MyBox[2]{
  \fbox{\lower0.75cm
    \vbox to 1.7cm{\vfil
      \hbox to 1.7cm{\hfil\parbox{1.4cm}{#1\\#2}\hfil}
      \vfil}%
  }%
}

\newcommand{\Defset}[3][\defsep]{\Set{#2#1#3}}

\newcommand{\Set}[1]{\left\{#1\right\}}

\newcommand{\field}{\mathbb}

\newcommand{\reals}{\field{R}}

\newcommand{\R}{\reals}

\DeclareMathOperator*{\argmin}{arg\,min}

\newcommand{\norm}[2][]{\lVert\fcarg{#2}\rVert\ifx#1\empty\else_{#1}\fi}
\newcommand{\Norm}[2][]{\left\lVert#2\right\rVert\ifx#1\empty\else_{#1}\fi}

\newcommand{\AC}{\text{AC}\xspace}
\newcommand{\TP}{\text{TP}\xspace}
\newcommand{\TN}{\text{TN}\xspace}
\newcommand{\FP}{\text{FP}\xspace}
\newcommand{\FN}{\text{FN}\xspace}

\newcommand{\MCC}{\text{MCC}\xspace}

\newcommand{\codename}[1]{\textsf{#1}}

\makeatletter
\newcommand{\fcdot}{\,\cdot\,}
\newcommand{\fcarg}[1]{\def\fc@rg{#1}\ifx\fc@rg\empty\fcdot\else\fc@rg\fi}
\makeatother

\bibliography{poisoning_attacks_LR}

\begin{document}

\title[Poisoning Attacks on Logistic Regression]%
{A Convex-Analytic Reformulation for\\Poisoning Attacks on Logistic Regression}

\author[ C. M. Costa, M. E. Pinheiro, M. Schmidt]%
{Carina Moreira Costa, Maria Eduarda Pinheiro, Martin Schmidt}

\address[C. M. Costa]{%
State University of Maringá,
Department of Mathematics,
Maringá,
Brazil}
\email{cmcosta@uem.br}

\address[M. E. Pinheiro] {%
Federal University of Santa Catarina,
Department of Mathematics,
Blumenau Campus,
Blumenau,
Brazil}
\email{maria.eduarda.pinheiro@ufsc.br}

\address[M. Schmidt]{%
  Trier University,
  Department of Mathematics,
  Universitätsring 15,
  54296 Trier,
  Germany}
\email{martin.schmidt@uni-trier.de}


\begin{abstract}
  We study data poisoning attacks on logistic regression with
$\ell_2$-regularization, where an attacker chooses the features of a
fixed number of injected observations with prescribed labels to
maximize the logistic loss on the clean training data. Strong
convexity of the defender's problem yields an exact reformulation
through its first-order optimality conditions. Using convex conjugacy
of the softplus function, we replace the sigmoid relations by a
single convex optimality-gap inequality and express the attacker's
objective as a concave function. A change of variables makes the
stationarity condition affine, leaving bounded bilinear equalities
as the only nonconvex constraints. We derive finite bounds on all
variables involved in these equalities, including a global bound on
the coefficient norm and tighter bounds obtained by preprocessing
based on convex optimization. Experiments on eight binary
classification datasets evaluate attack effectiveness and compare the
two reformulations when solved with \textsf{SCIP}.
Attacks obtained by local optimization degrade
predictive performance on most datasets, while the convex-analytic
formulation produces substantially stronger reported dual bounds
within a fixed time limit.


\end{abstract}

\keywords{Data poisoning,
Logistic regression,
Bilevel optimization,
Convex conjugacy,
Global optimization,
Bound tightening%
}
\subjclass[2020]{Primary 90C26; Secondary 90C25, 91A65, 62J12%
}

\maketitle

\section{Introduction}
\label{sec:introduction}

Machine learning models support automated decision-making
across many fields and rely on training data,
which can come from various sources, including the internet, humans, government
databases, third-party sources, among others. A major concern regarding this
is that data manipulations such as injection of incorrect or biased data into the
training set can have severe consequences; e.g., the compromised model may develop
security vulnerabilities, generate systematically biased or inaccurate predictions,
and ultimately fail when used in critical applications, such as fraud detection,
healthcare, etc.

One type of attack that a machine learning model can be subject to is the
so-called \textit{poisoning attack}. They occur when a malicious actor injects
specially crafted attack data samples into the training data, forcing the
model to learn erroneous patterns and degrading the predictive performance
of the model on unseen data. In this work, we consider poisoning attacks against
a logistic regression model. This attack is modeled as a bilevel optimization
problem representing a game between an attacker and a defender. In this
adversarial process, the attacker generates a set of poisoning points and the
learner tries to learn the best predictive regression model.

The use of mathematical optimization to study machine learning models,
including adversarial learning and machine teaching, is surveyed by
\textcite{gambella2021optimization}.
Adversarial manipulation of statistical inference is also studied by
\textcite{caballero2025indiscriminate}, who formulate attacks on
conditional inference under multivariate Gaussian models and exploit
their structure to develop solution methods. Their attacks modify
the evidence supplied to a fixed probabilistic model. In this paper,
the attacker instead injects training observations and thereby
influences the parameters learned by the defender.

The formulation of poisoning attacks against some machine learning models
as a bilevel optimization problem has already been considered in the literature.
Poisoning attacks against support vector machines (SVMs) are studied by
\textcite{biggio2012poisoning}. By differentiating the KKT conditions of SVM,
the method proposed in the paper iteratively computes the validation
loss gradient to optimize the attack point, in order to maximize the test
classification error. \textcite{xiao2015feature} explore the
vulnerability of some feature selection methods, including LASSO,
ridge regression, and elastic net, to training data poisoning.
The attacker aims to increase classification error by
maximizing the squared loss on the clean training or surrogate data plus a
regularization term. A gradient-ascent strategy is used. Their experiments on PDF malware
detection demonstrate that feature selection methods can be severely
compromised under attack.

Poisoning attacks in logistic regression have also been considered, but
the approach is different from the one we employ in this paper.
In \textcite{mei2015using}, the idea was to mathematically formalize how
an attacker can make the smallest possible modifications to the training
set to force the algorithm to learn a beneficial target model for the attacker.
The authors applied the idea to SVM, logistic regression, and linear regression.
The authors derive single-level reformulations using KKT
conditions and employ gradient descent for their attacks against SVMs
and logistic regression. In the particular case of
logistic regression, this framework was tested on a single dataset,
the Spambase data set, and the experiment consisted of generating
artificial target weights~$\beta^*$ to show that the optimization procedure
was able to find a poisoned data set whose learned weights
were close to~$\beta^*$. Global optimality of the computed attack
was not certified.

\textcite{carnerero2021regularization} consider logistic regression
with $\ell_{2}$-regularization and the attack is modeled as a minimax
bilevel optimization problem. To solve the problem, the authors employ a
hypergradient based approach. Unlike previous approaches, in which the
regularization parameter is assumed to remain fixed during an attack,
the authors sought to model an optimal attack where both the poisoned
data and the hyperparameters adjust dynamically under worst-case conditions.
The experiments show that as the number of poisoning points increases, the
learned regularization hyperparameter naturally increases to compensate for
the instability the attacker attempts to introduce, but, on the other hand,
it then forces the norm of the weight vector~$\beta$ to be closer to zero.

\textcite{steinhardt2017certified} derive minimax upper
bounds and candidate attacks for sanitized empirical risk minimization
over a constrained parameter set. Their surrogate uses the minimized
contaminated loss; its test-loss interpretation additionally relies on
train-test concentration and an approximation retaining all clean points
during sanitization. Here, we maximize exactly the clean training loss
under fixed poisoning labels and a quadratic defender penalty, without
sanitization.

\textcite{koh2022stronger}, first circulated in 2018 and
revised in 2021, develop influence, KKT, and minimax attacks, including
logistic regression. Their KKT attack selects decoy parameters and then
approximately enforces their optimality, rather than globally optimizing
clean loss over defender parameters. Their two-point concentration theorem
allows fractional copies and at most the original poisoning mass. It does
not establish preservation of our prescribed integer class counts and
exactly $n+q$ observations, so it does not by itself justify restricting
our model to two poisoning locations.

\textcite{suya2021model} construct model-targeted attacks
with convergence to attainable targets under their assumptions and a lower
bound on the number of poisoning points needed to reach a target. These
guarantees concern target attainment as the budget grows, rather than
global clean-loss maximization at a fixed budget.

More recently, \textcite{schewe2025ridge} and \textcite{csuvak2026design}
have studied poisoning attacks against ridge regression and linear
regression, respectively. In the upper-level problem of
\textcite{csuvak2026design}, the attacker selects the features of the
poisoning samples to maximize the MSE of the regression model on the true
training data. In the lower-level problem, the learner simultaneously
optimizes its parameters, weights and bias, to minimize its loss function
(with an~$\ell_{2}$-norm regularization term) over the  poisoned training
dataset. The experiments show that the proposed attack strategies
are superior to a benchmark algorithm from the literature.
\textcite{schewe2025ridge} propose a novel approach for poisoning attacks
against ridge regression in which both numerical and categorical features
are considered.
The attack is framed as a mixed-integer bilevel optimization problem,
where categorical variables are modeled as binary decisions
that belong to SOS-1 sets.
To solve the problem, it is first transformed into a single-level
mixed-integer nonlinear optimization problem.
A well-known computational bottleneck in single-level KKT reformulations
is that they introduce lower-level variables that have no bounds.
When solvers cannot bound these variables, their performance can
be very poor. Thus, to improve solver performance, a method for
bounding the lower-level variables is presented. However,
even with tighter bounds, only toy examples could be solved
to global optimality.
The authors then propose a heuristic approach to quickly compute
high-quality feasible solutions to the single-level reformulation.
Finally, the authors emphasize the need to explore more robust exact
approaches in the context of adversarial machine learning.

In this paper, we study poisoning attacks against
$\ell_2$-regularized logistic regression from a global optimization
point of view. We first derive a single-level formulation of the
bilevel problem using the optimality conditions of the defender's
problem. KKT replacement, the softplus conjugate, and
McCormick envelopes~\parencite{mccormick1976computability} are established
ingredients. Our contribution is their integration into an exact
formulation of the specified fixed-label, fixed-budget problem that
combines a concave representation of the clean-loss objective, affine
stationarity, bounded bilinear equalities as the only nonconvex
constraints, and computable valid variable bounds. To this end,
we use the convex conjugate of the softplus function to replace the
sigmoid relations by a single convex inequality and introduce a
change of variables that renders the stationarity condition affine.
Moreover, we derive valid bounds on the classifier coefficients and
the auxiliary variables, including tighter bounds obtained by convex
preprocessing. These bounds allow us to construct convex relaxations
of the remaining bilinear terms for spatial branch-and-bound. In our
numerical study on eight binary classification datasets, we evaluate
the effect of the attacks on predictive performance and compare the
two single-level formulations using \textsf{SCIP}. The results show
substantial improvements in the dual bounds reported by \textsf{SCIP}, while
also illustrating the difficulty of certifying global optimality.

The remainder of the paper is structured as follows.
In Section~\ref{sec:model}, we present the bilevel poisoning model
and its reformulation using the first-order optimality conditions
of the lower-level problem. The convex-analytic reformulation,
the valid variable bounds, and the resulting convexity structure
are discussed in Section~\ref{sec:reformulation}.
In Section~\ref{sec:numerical_results}, we present and discuss our
numerical results. Finally, we close with some concluding remarks in
Section~\ref{sec:conclusion}.


\section{Model}
\label{sec:model}

We consider a binary classification setting in which a defender trains a
logistic regression model on a dataset that may have been partially
corrupted by an adversary. The attacker's goal is to degrade the
classifier's performance on clean data by injecting a small number of
carefully crafted poisoning points into the training set.

Let $\mathcal{D}^{\mathrm{tr}} = \{(x_i^{\mathrm{tr}},\,
y_i^{\mathrm{tr}})\}_{i=1}^{n}$ be the training set, with labels
$y_i^{\mathrm{tr}} \in \{0,1\}$ and features $x_i^{\mathrm{tr}} \in
[0,1]^d$.
We assume throughout that all features have been scaled to $[0,1]$
as part of the preprocessing. The regularization term below is defined
with respect to these scaled features.
We fix $x_{i,1}^{\mathrm{tr}} = 1$ for all~$i$, so that the intercept
is considered in the coefficient vector.
A point $x_i^{\mathrm{tr}}$ is said to belong to the \emph{positive
  class} if $y_i^{\mathrm{tr}} = 1$, and to the \emph{negative class}
otherwise.

The logistic regression classifier assigns to each point $x_i$ the
probability
\begin{equation}\label{probability}
  m_\beta(x_i)
  =
  \frac{1}{1 + e^{-\beta^\top x_i}} \in (0,1)
\end{equation}
of belonging to the positive class,
and the coefficients~$\beta$ are obtained by
minimizing the negative log-likelihood
\begin{equation}\label{LR}
  \min_{\beta\in \R^d}\;
  \mathcal{L}(\mathcal{D}^{tr}, \beta)
  + \lambda \|\beta\|_2^2,
\end{equation}
where $\lambda > 0$ is a ridge penalty and
\[
  \mathcal{L}(\mathcal{D}^{tr}, \beta)
  \define
  -\sum_{i=1}^{n}
  \bigl[
    y_i^{tr} \log m_\beta(x_i^{tr})
    +
    (1-y_i^{tr})\log \left(1 - m_\beta(x_i^{tr})\right)
  \bigr].
\]

The regularization term ensures coercivity of the lower-level
objective and, as shown below, makes it $2\lambda$-strongly
convex. Consequently, the lower-level problem admits a unique
minimizer.

\subsection{Poisoning Attack}

The attacker constructs a set of poisoning points
\[
  \mathcal{D}^{p} = \{(x_k^{p},\, y_k^{p})\}_{k=1}^{q},
  \quad
  y_k^{p} \in \{0,1\} \;\text{fixed},
  \quad
  x_k^{p} \in [0,1]^d,
\]
whose labels are chosen in advance and whose features are the attacker's
decision variables.  As for the training data, we fix~$x_{k,1}^p = 1$
for all~$k$, so that the intercept convention is maintained. The
attacker then injects $\mathcal{D}^{p}$ into
the training set, so the defender unknowingly solves~\eqref{LR} on the
contaminated data $\mathcal{D}^{tr} \cup \mathcal{D}^{p}$.
Here and throughout, $\mathcal{D}^{tr} \cup \mathcal{D}^{p}$ denotes
the combined indexed dataset of $n+q$ observations. Repeated
feature-label pairs are retained and contribute separately to the loss.

The attacker's objective is to maximize the loss that the resulting
classifier incurs on the clean training data.  This yields the
bilevel optimization problem
\begin{varsubequations}{P1}\label{PA}
  \begin{align}
    \max_{x_k^{p}}\quad
      & \mathcal{L}(\mathcal{D}^{tr},\, \hat{\beta})
        \label{eq:P1_obj} \\
    \st\quad
      & x_{k,1}^p = 1,
        \quad x_{k,j}^p \in [0,1],
        \quad j = 2,\dots,d, \quad
        \quad k = 1,\dots,q\,
        \label{eq:P1_box} \\
      & \hat{\beta}
        =
        \argmin_{\beta \in \R^d} \Set{
        \mathcal{L}(\mathcal{D}^{tr} \cup \mathcal{D}^{p},\, \beta)
        + \lambda\|\beta\|_2^2 },
        \label{min_beta}
  \end{align}
\end{varsubequations}
where
\begin{equation}
  \label{lossfunctiontrp}
  \begin{split}
    \mathcal{L}(\mathcal{D}^{tr} \cup \mathcal{D}^{p}, \beta)
    = &
    - \sum_{i=1}^{n}
    \bigl[
    y_i^{tr} \log m_\beta(x_i^{tr})
    + (1-y_i^{tr})\log\!\bigl(1 - m_\beta(x_i^{tr})\bigr)
    \bigr]\\
    & - \sum_{k=1}^{q}
    \bigl[
    y_k^p \log m_\beta(x_k^{p})
    + (1-y_k^p)\log\!\bigl(1 - m_\beta(x_k^{p})\bigr)
    \bigr].
  \end{split}
\end{equation}

Note that the upper-level objective~\eqref{eq:P1_obj} measures the
classifier's logistic loss on the clean data only, whereas the
lower-level problem~\eqref{min_beta} trains on the contaminated set.  The
regularization appears only in the lower level and it reflects the
defender's training procedure, not the attacker's success criterion.

\subsection{Gradient and Hessian of the Lower-Level Objective Function}

To reformulate the bilevel problem as a single-level one, we need the
first-order optimality conditions of the lower level.  Let
$\mathcal{L}(\beta) \define \mathcal{L}(\mathcal{D}^{tr} \cup
\mathcal{D}^{p}, \beta) + \lambda\|\beta\|_2^2$.
Stacking training and poisoning data, we define
\begin{equation*}
  X =
  \begin{bmatrix}
    x_1^{tr\top} \\
    \vdots \\
    x_n^{tr\top} \\
    x_1^{p\top} \\
    \vdots \\
    x_q^{p\top}
  \end{bmatrix}
  \in \R^{(n+q)\times d},
  \quad
  y =
  \begin{pmatrix}
    y_1^{tr} \\
    \vdots \\
    y_n^{tr} \\
    y_1^{p} \\
    \vdots \\
    y_q^{p}
  \end{pmatrix}
  \in \R^{n+q},
  \quad
  m_\beta(X) =
  \begin{pmatrix}
    m_\beta(x_1^{tr}) \\
    \vdots \\
    m_\beta(x_n^{tr}) \\
    m_\beta(x_1^{p}) \\
    \vdots \\
    m_\beta(x_q^{p})
  \end{pmatrix}
  \in \R^{n+q}.
\end{equation*}
In this notation, the regularized lower-level objective is
\[
  \mathcal{L}(\beta)
  =
  -y^\top \log m_\beta(X)
  -(\mathbf{1}-y)^\top \log\!\bigl(\mathbf{1}-m_\beta(X)\bigr)
  +\lambda\|\beta\|_2^2,
\]
where the logarithm is applied componentwise and
$\mathbf{1}\in\R^{n+q}$ denotes the vector of all ones.
The gradient and Hessian are then given by
\begin{equation}\label{eq:grad_hess}
  \nabla\mathcal{L}(\beta)
  = X^\top\left(m_\beta(X) - y\right) + 2\lambda\beta,
  \quad
  \nabla^2\mathcal{L}(\beta)
  = X^\top \Gamma\, X + 2\lambda I,
\end{equation}
where
$\Gamma = \operatorname{diag}(\gamma_1, \dots, \gamma_{n+q})$
with
\begin{equation}\label{gamma}
  \gamma_i =
  \begin{cases}
    m_\beta(x_i^{tr})\left(1 - m_\beta(x_i^{tr})\right),
      & 1 \leq i \leq n, \\[4pt]
    m_\beta(x_{i-n}^{p})\left(1 - m_\beta(x_{i-n}^{p})\right),
      & n{+}1 \leq i \leq n{+}q.
  \end{cases}
\end{equation}

\subsection{Strong Convexity and Single-Level Reformulation}

\begin{prop}
  \label{prop:strong-convexity}
  For every $\lambda > 0$, the regularized loss
  \[
    \mathcal{L}(\mathcal{D}^{\mathrm{tr}} \cup \mathcal{D}^{\mathrm{p}},\beta)
    + \lambda \norm{\beta}_2^2
  \]
  is $2\lambda$-strongly convex in $\beta$.
\end{prop}

\begin{proof}
  By~\eqref{eq:grad_hess}, the Hessian of the regularized loss is
  \[
    \nabla^2 \mathcal{L}(\beta)
    =
    X^\top \Gamma X + 2\lambda I.
  \]
  Since each diagonal entry of $\Gamma$ is of the form
  \[
    \gamma_i
    =
    m_\beta(x_i)\bigl(1-m_\beta(x_i)\bigr)
    \geq 0,
  \]
  the matrix $\Gamma$ is positive semidefinite. Hence, for every
  $v \in \R^d$,
  \[
    v^\top \nabla^2 \mathcal{L}(\beta) v
    =
    (Xv)^\top \Gamma (Xv)
    + 2\lambda \norm{v}_2^2
    \geq
    2\lambda \norm{v}_2^2.
  \]
  Therefore,
  \[
    \nabla^2 \mathcal{L}(\beta) \succeq 2\lambda I
  \]
  for every $\beta \in \R^d$. Thus, $\mathcal{L}$ is
  $2\lambda$-strongly convex.
\end{proof}

Moreover, the lower-level objective is continuous and coercive, since
\[
  \mathcal{L}(\mathcal{D}^{\mathrm{tr}} \cup \mathcal{D}^{\mathrm{p}},\beta)
  + \lambda \norm{\beta}_2^2
  \geq
  \lambda \norm{\beta}_2^2
  \to \infty
  \quad\text{as}\quad\norm{\beta}_2 \to \infty.
\]
Hence, the lower-level problem~\eqref{min_beta} admits a minimizer.
By Proposition~\ref{prop:strong-convexity}, this minimizer is unique and we
denote it by~$\hat\beta$. Since the lower-level objective is
differentiable and strongly convex and the problem is unconstrained,
$\hat\beta$ is characterized by the first-order optimality condition
\[
  \nabla \mathcal{L}(\hat\beta) = 0.
\]
Hence, we obtain the single-level reformulation
\begin{varsubequations}{P2}\label{PA_2}
  \begin{align}
    \max_{x_k^{p},\, \beta}\quad
      & \mathcal{L}(\mathcal{D}^{tr}, \beta)
        \label{eq:P2_obj} \\
      \st\quad
      & x_{k,1}^p = 1,
        \quad x_{k,j}^p \in [0,1],
        \quad j =  2,\dots,d,
        \quad k = 1,\dots,q,
        \label{eq:P2_box} \\
      & X^\top\!\bigl(m_\beta(X) - y\bigr) + 2\lambda\beta = 0.
        \label{eq:P2_kkt}
  \end{align}
\end{varsubequations}


\section{Convex-Analytic Reformulation}
\label{sec:reformulation}

The single-level formulation~\eqref{PA_2} is nonconvex
since the stationarity constraint~\eqref{eq:P2_kkt} contains
sigmoid evaluations and the poisoning
features~$x_k^p$ interact bilinearly with the
classifier coefficients~$\beta$.
In this section, we derive an equivalent
reformulation in which the only nonconvexities are
bounded bilinear terms.

We begin in Section~\ref{sec:signed-margins} by
rewriting the logistic loss using signed margins and  softplus
and by establishing the
convex-conjugacy properties of the softplus.
In Section~\ref{sec:convex-reformulation}, we use these
properties to replace the nonconvex sigmoid evaluations
in the stationarity condition by a convex Fenchel-gap
inequality, and rewrite the upper-level objective in
concave form.
Section~\ref{sec:eliminating-xp} eliminates the
poisoning features~$x_k^p$ by a change of variables
that renders the stationarity condition linear.
Section~\ref{subsec:valid-bounds} derives finite bounds
on all variables from the problem data and the
regularization parameter, without arbitrary big-$M$
constants.
The resulting reformulation~\eqref{PA_3} and its
equivalence with~\eqref{PA_2} are presented in
Section~\ref{sec:final-reformulation}.

\subsection{Signed Margins and Convex Conjugacy}
\label{sec:signed-margins}

For a label $y\in\{0,1\}$, define
\begin{equation}
  \label{eq:a_definition}
  a \define 1-2y \in \{-1,1\}.
\end{equation}

Accordingly, we set
\[
  s_i^{tr} \define 1-2y_i^{tr},
  \quad
  s_k^p \define 1-2y_k^p.
\]
Moreover, let
\begin{equation}
  \label{eq:softplus}
  \phi(\theta)
  \define
  \log(1+e^\theta),
  \quad \theta\in\R.
\end{equation}
For every observation $(x,y)$, its logistic-loss contribution can then
be written as
\begin{equation}
  \label{eq:signed-loss}
  -y\log m_\beta(x)
  -(1-y)\log\bigl(1-m_\beta(x)\bigr)
  =
  \phi\bigl((1-2y)\beta^\top x\bigr).
\end{equation}
To see this, note that for $y=0$ the left-hand side reduces to
$$-\log(1-m_\beta(x)) = \log(1+e^{\beta^\top x})= \phi(\beta^\top x).$$
For $y=1$ it reduces to
$-\log m_\beta(x) = \log(1+e^{-\beta^\top x})$,
which equals $\phi(-\beta^\top x)$.
In both cases the right-hand side of~\eqref{eq:signed-loss} matches.
We therefore introduce signed-margin variables
\begin{equation}
  \label{eq:theta-tr}
  \theta_i^{tr}
  =
  s_i^{tr}\beta^\top x_i^{tr},
  \quad i=1,\dots,n.
\end{equation}
For the poisoning points, corresponding variables~$\theta_k^p$ will
be introduced below.
The derivative of~$\phi$ is the sigmoid
\[
  \phi'(\theta)
  =
  \frac{1}{1+e^{-\theta}}.
\]
The following convex-conjugacy relation will allow us to avoid
introducing a nonconvex sigmoid equality.

\begin{prop}
  \label{prop:logistic-conjugate}
  The convex conjugate of~$\phi$ is
  \begin{equation}
    \label{eq:phi-conjugate}
    \phi^*(\alpha)
    =
    \alpha\log\alpha
    +(1-\alpha)\log(1-\alpha),
    \quad
    \alpha\in[0,1],
  \end{equation}
  where $0\log 0$ is interpreted as~$0$.
  Moreover, for every $\theta\in\R$ and $\alpha\in[0,1]$,
  \begin{equation}
    \label{eq:fenchel}
    \phi(\theta)+\phi^*(\alpha)-\alpha\theta
    \geq 0,
  \end{equation}
  and equality holds if and only if
  \begin{equation}
    \label{eq:alpha-sigmoid}
    \alpha
    =
    \phi'(\theta)
    =
    \frac{1}{1+e^{-\theta}}.
  \end{equation}
\end{prop}

\begin{proof}
  By definition,
  \[
    \phi^*(\alpha)
    =
    \sup_{\theta\in\R}
    \bigl\{
    \alpha\theta - \log(1+e^\theta)
    \bigr\}.
  \]
  For $\alpha\in(0,1)$, define
  $g(\theta) := \alpha\theta - \log(1+e^\theta)$.
  Its first and second derivatives are
  \[
    g'(\theta)
    =
    \alpha - \frac{e^\theta}{1+e^\theta},
    \quad
    g''(\theta)
    =
    -\frac{e^\theta}{(1+e^\theta)^2}
    < 0
    \quad\text{for all } \theta\in\R.
  \]
  Hence, $g$ is strictly concave, so any stationary point is the
  unique global maximizer and satisfies
  \begin{equation}\label{eq:alpha-sigmoid_2}
    \alpha
    =
    \frac{e^\theta}{1+e^\theta},
  \end{equation}
  which is equivalent to
  \[
    \theta
    =
    \log\frac{\alpha}{1-\alpha}.
  \]
  Since~$g$ is strictly concave and~$\theta^*$ is its
  unique maximizer,
  \[
    \phi^*(\alpha)
    =
    \sup_{\theta \in \R} g(\theta)
    =
    g(\theta^*).
  \]
  Evaluating~$g(\theta^*)$ with
  $\theta^* = \log\frac{\alpha}{1-\alpha}$ gives
  \begin{align*}
    g(\theta^*)
    &=
      \alpha\log\frac{\alpha}{1-\alpha}
      -
      \log\!\left(1+\frac{\alpha}{1-\alpha}\right)
    \\
    &=
      \alpha\log\frac{\alpha}{1-\alpha}
      -
      \log\frac{1}{1-\alpha}
    \\
    &=
      \alpha\log\alpha - \alpha\log(1-\alpha)
      +
      \log(1-\alpha)
    \\
    &=
      \alpha\log\alpha + (1-\alpha)\log(1-\alpha),
  \end{align*}
  which proves~\eqref{eq:phi-conjugate}.
  The values at $\alpha=0$ and $\alpha=1$ follow by continuity.

  Inequality~\eqref{eq:fenchel} is Fenchel's inequality, which
  holds for any function and its convex conjugate.  Since $\phi$
  is differentiable and strictly convex, equality in Fenchel's
  inequality holds if and only if $\alpha = \phi'(\theta)$, which
  gives~\eqref{eq:alpha-sigmoid}.
\end{proof}

\subsection{Reformulation of the Stationarity Condition
  and Upper-Level Objective}
\label{sec:convex-reformulation}

\subsubsection{Convex Optimality-Gap Constraint}
\label{sec:convex-gap}
The stationarity condition of the lower-level problem involves
evaluations of $1/(1+e^{-\theta})$,
which is a nonconvex function of~$\theta$.  Our goal in this
section is to replace the stationarity condition by a system of
constraints that is equivalent but whose only nonconvexities are
bilinear terms.

For the moment, suppose that signed
margins~$\theta_i^{tr}$ and~$\theta_k^p$ are available for all
training and poisoning points.
Using~\eqref{eq:signed-loss}, each loss contribution
can be written as~$\phi(\theta)$, so the stationarity
condition of the lower-level problem reads
\begin{equation}
  \label{eq:stationarity-phi}
  \sum_{i=1}^n
  s_i^{tr}\,\phi'(\theta_i^{tr})\,x_i^{tr}
  +
  \sum_{k=1}^q
  s_k^p\,\phi'(\theta_k^p)\,x_k^p
  +
  2\lambda\beta
  =
  0.
\end{equation}
This condition is nonconvex because of the
evaluations of~$\phi'$.
We add new variables
\begin{equation}
  \label{eq:alpha-bounds}
  \alpha_i^{tr}\in[0,1],\quad i=1,\ldots,n,
  \quad
  \alpha_k^p\in[0,1],\quad k=1,\ldots,q,
\end{equation}
to the problem and consider the bilinear condition
\begin{equation}
  \label{eq:stationarity-alpha-explicit}
  \sum_{i=1}^n
  s_i^{tr}\,x_i^{tr}\,\alpha_i^{tr}
  +
  \sum_{k=1}^q
  s_k^p\,x_k^p\,\alpha_k^p
  +
  2\lambda\beta
  =
  0,
\end{equation}
which is obtained from~\eqref{eq:stationarity-phi} by writing
$\alpha_i$ in place of~$\phi'(\theta_i)$.
Every point satisfying~\eqref{eq:stationarity-phi} also
satisfies~\eqref{eq:stationarity-alpha-explicit}, by
setting~$\alpha_i = \phi'(\theta_i)\in(0,1)$, but the
converse does not hold because
Condition~\eqref{eq:stationarity-alpha-explicit} accepts
any~$\alpha\in[0,1]$ and is therefore a relaxation of the
original stationarity condition.

We now show that adding a single convex inequality restores
the equivalence.
Taking the scalar product
of~\eqref{eq:stationarity-alpha-explicit} with~$\beta$ and
using~$\theta_i^{tr} = s_i^{tr}\beta^\top x_i^{tr}$
and~$\theta_k^p = s_k^p\beta^\top x_k^p$ gives
\begin{equation}
  \label{eq:stationarity-dot-beta}
  \sum_{i=1}^n
  \alpha_i^{tr}\,\theta_i^{tr}
  +
  \sum_{k=1}^q
  \alpha_k^p\,\theta_k^p
  +
  2\lambda\|\beta\|_2^2
  =
  0.
\end{equation}
It remains to recover the equivalence with the original
stationarity condition~\eqref{eq:stationarity-phi}, i.e., to ensure
that~$\alpha_i = \phi'(\theta_i)$ holds at every feasible point.
Recall from Proposition~\ref{prop:logistic-conjugate}
that~$\alpha = \phi'(\theta)$ if and only if the Fenchel
gap~\eqref{eq:fenchel} vanishes.
The sum of all individual Fenchel gaps can be written as
\begin{align}
  &
    \sum_{i=1}^n
    \left[
    \phi(\theta_i^{tr})
    +\phi^*(\alpha_i^{tr})
    -\alpha_i^{tr}\,\theta_i^{tr}
    \right]
    +
    \sum_{k=1}^q
    \left[
    \phi(\theta_k^p)
    +\phi^*(\alpha_k^p)
    -\alpha_k^p\,\theta_k^p
    \right]
    \nonumber\\
  =\ &
    \sum_{i=1}^n
    \left[
    \phi(\theta_i^{tr})
    +\phi^*(\alpha_i^{tr})
    \right]
    +
    \sum_{k=1}^q
    \left[
    \phi(\theta_k^p)
    +\phi^*(\alpha_k^p)
    \right]
    +
    2\lambda\|\beta\|_2^2,
    \label{eq:sum-fenchel-gaps}
\end{align}
where the equality follows
from~\eqref{eq:stationarity-dot-beta}.
By Proposition~\ref{prop:logistic-conjugate}, every individual
Fenchel gap is nonnegative, so the right-hand side
in~\eqref{eq:sum-fenchel-gaps} satisfies
\begin{equation}
  \label{eq:convex-gap}
  \sum_{i=1}^n
  \left[
    \phi(\theta_i^{tr})
    +\phi^*(\alpha_i^{tr})
  \right]
  +
  \sum_{k=1}^q
  \left[
    \phi(\theta_k^p)
    +\phi^*(\alpha_k^p)
  \right]
  +
  2\lambda\|\beta\|_2^2
  \geq 0.
\end{equation}
The left-hand side of~\eqref{eq:convex-gap} is convex,
since~$\phi$, $\phi^*$, and~$\|\cdot\|_2^2$ are all convex.
Now, we impose the reverse inequality as  a constraint:
\begin{equation}
  \label{eq:convex-gap-leq}
  \sum_{i=1}^n
  \left[
    \phi(\theta_i^{tr})
    +\phi^*(\alpha_i^{tr})
  \right]
  +
  \sum_{k=1}^q
  \left[
    \phi(\theta_k^p)
    +\phi^*(\alpha_k^p)
  \right]
  +
  2\lambda\|\beta\|_2^2
  \leq 0.
\end{equation}
Together, \eqref{eq:convex-gap} and~\eqref{eq:convex-gap-leq}
force the sum to be zero.  Since every individual Fenchel gap
is nonnegative and their sum is zero, each gap must vanish.
By~\eqref{eq:alpha-sigmoid} in
Proposition~\ref{prop:logistic-conjugate}, this
implies~$\alpha_i^{tr}=\phi'(\theta_i^{tr})$
and~$\alpha_k^p=\phi'(\theta_k^p)$.

Therefore, the original stationarity
condition~\eqref{eq:stationarity-phi} in the
variables~$(x^p,\beta)$ is equivalent to
conditions~\eqref{eq:alpha-bounds},
\eqref{eq:stationarity-alpha-explicit},
and~\eqref{eq:convex-gap-leq} in the extended
variables~$(x^p,\beta,\alpha)$: the bilinear
relaxation~\eqref{eq:stationarity-alpha-explicit} enlarges the
feasible set by introducing the~$\alpha$
variables~\eqref{eq:alpha-bounds}, and the convex
inequality~\eqref{eq:convex-gap-leq} compensates this exactly,
forcing~$\alpha_i = \phi'(\theta_i)$ at every feasible point.

\subsubsection{Concave Form of the Objective}
\label{subsec:concave-objective}

Finally, the upper-level objective can be rewritten in a concave way.
As shown above, Conditions~\eqref{eq:convex-gap}
and~\eqref{eq:convex-gap-leq} force all Fenchel
gaps~\eqref{eq:fenchel} to vanish.
In particular, for each $i = 1, \dots, n$,
\[
  \phi(\theta_i^{tr}) + \phi^*(\alpha_i^{tr})
  - \alpha_i^{tr}\,\theta_i^{tr} = 0
\]
holds. Summing over the training points gives
\begin{equation} \label{eq:loss_up-phi}
  \sum_{i=1}^n \phi(\theta_i^{tr})
  =
  \sum_{i=1}^n \alpha_i^{tr}\,\theta_i^{tr}
  -
  \sum_{i=1}^n \phi^*(\alpha_i^{tr}),
\end{equation}
and by~\eqref{eq:signed-loss} and \eqref{eq:theta-tr} the left-hand side
equals~$\mathcal{L}(\mathcal{D}^{tr},\beta)$.

Similarly, for each poisoning point,~$k = 1, \dots, q$, the
vanishing Fenchel gap gives
$\alpha_k^p\,\theta_k^p
= \phi(\theta_k^p)+\phi^*(\alpha_k^p)$.
Substituting these relations into \eqref{eq:stationarity-dot-beta} yields
\[
  \sum_{i=1}^n \alpha_i^{tr}\,\theta_i^{tr}
  =
  -\sum_{k=1}^q
  \bigl[\phi(\theta_k^p)+\phi^*(\alpha_k^p)\bigr]
  -
  2\lambda\|\beta\|_2^2.
\]
Therefore, using \eqref{eq:loss_up-phi}, we obtain
\begin{align}
  \mathcal{L}(\mathcal{D}^{tr},\beta)
  &=
    -
    \sum_{i=1}^n\phi^*(\alpha_i^{tr})
    -
    2\lambda\|\beta\|_2^2   -
    \sum_{k=1}^q
    \left[
    \phi(\theta_k^p)
    +\phi^*(\alpha_k^p)
    \right].
    \label{eq:concave-objective}
\end{align}
The right-hand side of~\eqref{eq:concave-objective} is concave
because $-\phi$, $-\phi^*$, and $-\|\cdot\|_2^2$ are concave.

\subsection{Eliminating the Poisoning Features}
\label{sec:eliminating-xp}

The stationarity condition~\eqref{eq:stationarity-alpha-explicit}
contains the products~$x_k^p\,\alpha_k^p$, which are bilinear in
the variables~$x^p$ and $\alpha$.  We now eliminate~$x_k^p$
by a change of variables that renders the stationarity condition
linear.

To this end, we define
\begin{equation}
  \label{eq:h-definition}
  h_k
  \define
  \alpha_k^p x_k^p
  \in\R^d,
  \quad k=1,\dots,q.
\end{equation}
For the intercept component~$j=1$, since~$x_{k,1}^p = 1$ holds,
\eqref{eq:h-definition} gives~$h_{k,1} = \alpha_k^p$,
so the bound~\eqref{eq:h-box} right below holds with equality for~$j=1$.

At every point satisfying~\eqref{eq:alpha-sigmoid}, we have
$\alpha_k^p\in(0,1)$.  Therefore,
$x_k^p\in[0,1]^d$ is equivalent to
\begin{equation}
  \label{eq:h-box}
  0
  \leq
  h_{k,j}
  \leq
  \alpha_k^p,
  \quad
  j=1,\dots,d.
\end{equation}
Moreover, since~$\alpha_k^p > 0$ holds at every feasible point,  multiplying
\[
  \theta_k^p = s_k^p\beta^\top x_k^p
\]
by~$\alpha_k^p$ and using~\eqref{eq:h-definition} yields the
equivalent relation
\begin{equation}
  \label{eq:reduced-margin}
  \alpha_k^p\theta_k^p
  =
  s_k^p\beta^\top h_k,
  \quad k=1,\dots,q.
\end{equation}
The poisoning part of stationarity becomes linear:
\[
  s_k^p x_k^p\alpha_k^p
  =
  s_k^p h_k.
\]
Hence, the complete stationarity condition
\eqref{eq:stationarity-alpha-explicit} is given by
\begin{equation}
  \label{eq:reduced-stationarity}
  \sum_{i=1}^n
  s_i^{tr}x_i^{tr}\alpha_i^{tr}
  +
  \sum_{k=1}^q
  s_k^p h_k
  +
  2\lambda\beta
  =
  0.
\end{equation}

\subsection{Valid Bounds}
\label{subsec:valid-bounds}

The remaining nonconvexities in the reformulation will be the
bilinear terms in~\eqref{eq:reduced-margin}.  Spatial
branch-and-bound therefore requires finite bounds on~$\beta$,
$\theta^p$, $\alpha^p$, and~$h$.  We next derive such bounds.

Let $N\define n+q$ and define
\[
  \sigma(z)
  \define
  \phi'(z)
  =
  \frac{1}{1+e^{-z}}.
\]

\subsubsection{A Global Bound on the Classifier Coefficients}

We first derive a stronger bound than the one obtained merely by
comparing the regularized objective value at~$\beta^*$ and at~$0$. For
this, we need an auxiliary lemma.
\begin{lem}
  \label{lem:H-rho-bound}
  Let~$\beta^*$ be the unique minimizer of the lower-level
  problem for an arbitrary admissible poisoning set.
  Define
  \begin{equation}
    \label{eq:H-R}
    H(r)
    \define
    2\lambda r^2
    +
    \sum_{i=1}^n
    \phi\!\left(-r\|x_i^{tr}\|_2\right)
    +
    q\,\phi(-r\sqrt d),
    \quad r\geq 0.
  \end{equation}
  Then, $H(\|\beta^*\|_2) \leq N\log 2$ holds.
\end{lem}

\begin{proof}
  Let
  \[
    F(\beta)
    \define
    \mathcal{L}
    (\mathcal{D}^{tr}\cup\mathcal{D}^p,\beta)
  \]
  denote the unregularized contaminated loss.
  At the lower-level minimizer~$\beta^*$,
  since~$\nabla \mathcal{L}(\beta^*)=0$ holds, we have
  \[
    \nabla F(\beta^*)
    =
    -2\lambda\beta^*.
  \]
  By~\eqref{eq:a_definition}--\eqref{eq:signed-loss}, it follows
  \begin{equation*}
    F(\beta)
    = \sum_{i=1}^n\phi(s_i^{tr}\beta^\top x_i^{tr})
    + \sum_{k=1}^q\phi(s_k^p\beta^\top x_k^p)
  \end{equation*}
  holds.
  Since~$\phi(0) = \log 2$, setting~$\beta=0$
  gives~$F(0)=N\log 2$.
  Because~$F$ is convex,
  \begin{equation*}
    N\log 2
    =
    F(0)
    \geq
    F(\beta^*)
    +
    \nabla F(\beta^*)^\top(0-\beta^*)
    =
    F(\beta^*)
    +
    2\lambda\|\beta^*\|_2^2.
  \end{equation*}
  Set~$\rho = \|\beta^*\|_2$.
  By~\eqref{eq:a_definition}, we get $|s_i^{tr}|=1$.
  The Cauchy--Schwarz inequality gives, for every clean
  observation,
  \[
    s_i^{tr}\beta^{*\top}x_i^{tr}
    \geq
    -\rho\,\|x_i^{tr}\|_2,
  \]
  and, because~$\phi$ is increasing, we obtain
  \[
    \phi
    \!\left(
      s_i^{tr}\beta^{*\top}x_i^{tr}
    \right)
    \geq
    \phi\!\left(-\rho\,\|x_i^{tr}\|_2\right).
  \]
  Similarly, for every poisoning point
  $x_k^p\in[0,1]^d$,
  \[
    s_k^p\beta^{*\top}x_k^p
    \geq
    -\rho\,\|x_k^p\|_2
    \geq
    -\rho\sqrt{d},
  \]
  so that
  $\phi(s_k^p\beta^{*\top}x_k^p)
  \geq \phi(-\rho\sqrt{d})$.
  Summing over all data points gives
  \begin{align*}
    F(\beta^*)
    &=
      \sum_{i=1}^n
      \phi\!\left(s_i^{tr}\beta^{*\top}x_i^{tr}\right)
      +
      \sum_{k=1}^q
      \phi\!\left(s_k^p\beta^{*\top}x_k^p\right)
    \\
    &\geq
      \sum_{i=1}^n
      \phi\!\left(-\rho\,\|x_i^{tr}\|_2\right)
      +
      q\,\phi(-\rho\sqrt{d})
    \\
    &=
      H(\rho) - 2\lambda\rho^2,
  \end{align*}
  where the last equality follows from the
  definition~\eqref{eq:H-R} of~$H$.
  Therefore, we finally get
  \[
    N\log 2
    \geq
    F(\beta^*) + 2\lambda\rho^2
    \geq
    H(\rho).
    \qedhere
  \]
\end{proof}
\begin{prop}
  \label{prop:beta-norm-bound}
  Let~$x^p$ be an arbitrary admissible poisoning set, i.e.,
  one satisfying~\eqref{eq:P1_box}, and let~$\beta^*$ be the
  corresponding unique minimizer of the lower-level
  problem~\eqref{min_beta}.  Consider~$H$ as in~\eqref{eq:H-R}
  and let~$R$ denote the largest nonnegative solution of
  \begin{equation}
    \label{eq:R-definition}
    H(R) = N\log 2.
  \end{equation}
  Then
  \begin{equation}
    \label{eq:beta-norm-R}
    \|\beta^*\|_2 \leq R.
  \end{equation}
  Moreover, $R < \sqrt{N\log 2 / (2\lambda)}$.
\end{prop}
\begin{proof}
  From~$\phi(0) = \log 2$, we obtain
  \[
    H(0)
    =
    0
    +
    \sum_{i=1}^n \phi(0)
    +
    q\,\phi(0)
    =
    N\log 2.
  \]
  Differentiating~\eqref{eq:H-R} at~$r=0$ gives
  \[
    H'(0)
    =
    -\phi'(0)
    \!\left(
      \sum_{i=1}^n \|x_i^{tr}\|_2
      +
      q\sqrt{d}
    \right)
    < 0,
  \]
  so~$H$ decreases below~$N\log 2$ for small
  positive~$r$.

  Since~$H$ is the sum of the strictly convex
  term~$2\lambda r^2$ and convex terms of the
  form~$\phi(-r\|x\|)$, it is strictly convex.
  Moreover $H(r) \to +\infty$ as~$r \to \infty$.
  Hence, the equation~$H(R) = N\log 2$ has a unique
  positive solution~$R$.
  The sublevel set
  \[
    \{r \geq 0 : H(r) \leq N\log 2\}
  \]
  is therefore the interval~$[0, R]$. By Lemma~\ref{lem:H-rho-bound},
  $H(\|\beta^*\|_2) \leq N\log 2$,
  so~$\|\beta^*\|_2 \leq R$.
  For the final claim, since~$\phi(\theta)
  = \log(1+e^\theta) > 0$ for all~$\theta \in \R$,
  the definition of~$H$ in~\eqref{eq:H-R} gives
  \[
    H(R)
    =
    2\lambda R^2
    +
    \underbrace{
      \sum_{i=1}^n
      \phi\!\left(-R\|x_i^{tr}\|_2\right)
      +
      q\,\phi(-R\sqrt{d})
    }_{> \; 0}
    >
    2\lambda R^2.
  \]
  Combining this with~$H(R) = N\log 2$ finally leads to
  $2\lambda R^2 < N\log 2$, i.e.,
  $R < \sqrt{N\log 2 / (2\lambda)}$.
\end{proof}
Proposition~\ref{prop:beta-norm-bound} is a statement about the
lower-level minimizer and does not depend on how the bilevel
problem is reformulated.  Every feasible point~$(x^p,\beta)$
of~\eqref{PA_2} satisfies the stationarity
condition~\eqref{eq:P2_kkt}, which, by
Proposition~\ref{prop:strong-convexity}, characterizes the
unique lower-level minimizer; hence $\beta = \beta^*$.
Consequently, the constraints
\[
  \beta \in [-R, R]^d,
  \quad
  |\beta^\top x| \leq R\,\|x\|_2 \leq R\sqrt{d}
  \quad \text{for } x \in [0,1]^d,
\]
can be added to~\eqref{PA_2} without excluding any feasible
point.  More generally, all bounds on~$\beta$, $\theta^{tr}$,
and~$\beta^\top x^p_k$  derived in this section are properties of the
lower-level minimizer and are therefore also valid
for~\eqref{PA_2}.
In our numerical experiments, both formulations are tested with the
bounds of Proposition~\ref{prop:beta-norm-bound} alone and with all
bounds of Section~\ref{subsec:valid-bounds}; see
Section~\ref{subsec:setup}.

\subsubsection{Componentwise Bounds}

The stationarity condition provides additional componentwise bounds.
Let
\[
  q_0
  \define
  \left|\Defset{k}{y_k^p=0}\right|,
  \quad
  q_1
  \define
  \left|\Defset{k}{y_k^p=1}\right|.
\]
For $j=1,\dots,d$, define
\begin{align}
  C_j^0
  &\define
    \sum_{\substack{i=1\\y_i^{tr}=0}}^n
  x_{i,j}^{tr}
  \sigma\!\left(R\|x_i^{tr}\|_2\right)
  +
  q_0\,\sigma(R\sqrt d),
  \label{eq:Cj0}\\
  C_j^1
  &\define
    \sum_{\substack{i=1\\y_i^{tr}=1}}^n
  x_{i,j}^{tr}
  \sigma\!\left(R\|x_i^{tr}\|_2\right)
  +
  q_1\,\sigma(R\sqrt d).
  \label{eq:Cj1}
\end{align}

\begin{prop}
  \label{prop:beta-component-bounds}
  Every lower-level minimizer~$\beta^*$ satisfies
  \begin{equation}
    \label{eq:beta-preliminary-bounds}
    \underline\beta_j^0
    \leq
    \beta_j^*
    \leq
    \overline\beta_j^0,
    \quad j=1,\dots,d,
  \end{equation}
  where
  \begin{equation}
    \label{eq:beta-preliminary-def}
    \underline\beta_j^0
    \define
    -\min\left\{
      R,\frac{C_j^0}{2\lambda}
    \right\},
    \quad
    \overline\beta_j^0
    \define
    \min\left\{
      R,\frac{C_j^1}{2\lambda}
    \right\}.
  \end{equation}
\end{prop}

\begin{proof}
  The $j$th component of the stationarity condition
  \eqref{eq:P2_kkt} can be written as
  \begin{align}
    2\lambda\beta_j^*
    &=\sum_{\substack{i=1\\y_i^{tr}=1}}^n
    x_{i,j}^{tr}
    \bigl(1-m_{\beta^*}(x_i^{tr})\bigr)+\sum_{\substack{k=1\\y_k^p=1}}^q
    x_{k,j}^p
    \bigl(1-m_{\beta^*}(x_k^p)\bigr)
    \nonumber\\
    & \quad
      -\sum_{\substack{i=1\\y_i^{tr}=0}}^n
    x_{i,j}^{tr}\,
    m_{\beta^*}(x_i^{tr})-\sum_{\substack{k=1\\y_k^p=0}}^q
    x_{k,j}^p\,
    m_{\beta^*}(x_k^p).
    \label{eq:stationarity-j}
  \end{align}

  Hence,
  \begin{equation}
    \label{eq:beta-j-sandwich}
    \begin{aligned}
      2\lambda\beta_j^*
      &\geq -
      \sum_{\substack{i=1\\y_i^{tr}=0}}^n
      x_{i,j}^{tr}\,
      m_{\beta^*}(x_i^{tr})
      -
      \sum_{\substack{k=1\\y_k^p=0}}^q
      x_{k,j}^p\,
      m_{\beta^*}(x_k^p)
      ,\\
      2\lambda\beta_j^*
      &\leq
      \sum_{\substack{i=1\\y_i^{tr}=1}}^n
      x_{i,j}^{tr}
      \bigl(1-m_{\beta^*}(x_i^{tr})\bigr)
      +
      \sum_{\substack{k=1\\y_k^p=1}}^q
      x_{k,j}^p
      \bigl(1-m_{\beta^*}(x_k^p)\bigr).
    \end{aligned}
  \end{equation}
  By Proposition~\ref{prop:beta-norm-bound},
  $|\beta^{*\top}x_i^{tr}| \leq R\|x_i^{tr}\|_2$ holds.
  Since~$\sigma$ is increasing,
  $m_{\beta^*}(x_i^{tr})
  = \sigma(\beta^{*\top}x_i^{tr})
  \leq \sigma(R\|x_i^{tr}\|_2)$. Moreover,
  $1~-~m_{\beta^*}(x_i^{tr})
  = \sigma(-\beta^{*\top}x_i^{tr})
  \leq \sigma(R\|x_i^{tr}\|_2)$.
  Analogously, for a poisoning point,
  \[
    |\beta^{*\top}x_k^p|
    \leq
    R\sqrt d,
  \]
  so both corresponding probabilities are bounded above by
  $\sigma(R\sqrt d)$.
  Substituting into the upper and lower bound
  of~\eqref{eq:beta-j-sandwich} and
  using~$x_{k,j}^p \leq 1$, gives
  \begin{align*}
    2\lambda\beta_j^*
    & \leq
      \sum_{\substack{i=1\\y_i^{tr}=1}}^n
    x_{i,j}^{tr}\,
    \sigma\!\left(R\|x_i^{tr}\|_2\right)
    +
    \sum_{\substack{k=1\\y_k^p=1}}^q
    1 \cdot \sigma(R\sqrt{d})
    =
      C_j^1,
    \\
  2\lambda\beta_j^*
  & \geq
    -\sum_{\substack{i=1\\y_i^{tr}=0}}^n
    x_{i,j}^{tr}\,
    \sigma\!\left(R\|x_i^{tr}\|_2\right)
    -
    q_0\,\sigma(R\sqrt{d})
    =
    -C_j^0.
  \end{align*}

  Therefore, we get
  \[
    -\frac{C_j^0}{2\lambda}
    \leq
    \beta_j^*
    \leq
    \frac{C_j^1}{2\lambda}.
  \]
  Intersecting these bounds with
  $|\beta_j^*|\leq\|\beta^*\|_2\leq R$
  gives~\eqref{eq:beta-preliminary-def}.
\end{proof}

The bounds in Proposition~\ref{prop:beta-component-bounds} are
closed-form bounds.  They can be tightened further by solving small
convex preprocessing problems. In fact, the proof of
Lemma~\ref{lem:H-rho-bound} shows that
\[
  N\log 2
  \geq
  F(\beta^*) + 2\lambda\|\beta^*\|_2^2.
\]
Splitting~$F(\beta^*)$ into its clean and poisoning
contributions and using the lower bound on the
poisoning part established in that proof gives
\begin{align*}
  F(\beta^*)
  &=  \sum_{i=1}^n \phi(s_i^{tr}\beta^{*\top}x_i^{tr})
    +
    \sum_{k=1}^q
    \phi(s_k^p\beta^{*\top}x_k^p)\\
  &\geq  \mathcal{L}(\mathcal{D}^{tr},\beta^*)  + q\phi(-\rho\sqrt{d}) \\
  & \geq \mathcal{L}(\mathcal{D}^{tr},\beta^*)  + q\,\phi(-R\sqrt{d}) \geq q\,\phi(-R\sqrt{d}).
\end{align*}
Substituting yields
\begin{equation}
  \label{eq:clean-valid-ineq}
  \mathcal{L}(\mathcal{D}^{tr},\beta^*)
  +
  2\lambda\|\beta^*\|_2^2
  \leq
  N\log 2
  -
  q\,\phi(-R\sqrt{d}).
\end{equation}
Hence, every feasible classifier coefficient vector belongs to the
compact convex set
\begin{equation}
  \label{eq:beta-outer-set}
  \mathcal{B}
  \define
  \left\{
    \beta\in\R^d
    \ \middle|\
    \begin{array}{l}
      \|\beta\|_2\leq R,\\[1mm]
      \underline\beta_j^0
      \leq\beta_j
      \leq\overline\beta_j^0,
      \quad j=1,\dots,d,\\[1mm]
      \mathcal{L}(\mathcal{D}^{tr},\beta)
      +2\lambda\|\beta\|_2^2
      \leq
      N\log2-q\,\phi(-R\sqrt d)
    \end{array}
  \right\}.
\end{equation}
We may therefore define the tighter bounds
\begin{equation}
  \label{eq:beta-obbt}
  \underline\beta_j
  \define
  \min_{\beta\in\mathcal{B}}\beta_j,
  \quad
  \overline\beta_j
  \define
  \max_{\beta\in\mathcal{B}}\beta_j.
\end{equation}
Both problems in~\eqref{eq:beta-obbt} are convex optimization
problems. The resulting bounds satisfy
$\underline\beta_j^0 \leq \underline\beta_j$ and
$\overline\beta_j \leq \overline\beta_j^0$,
that is, they are at least as tight as the closed-form
bounds~\eqref{eq:beta-preliminary-bounds}.
When solving~\eqref{eq:beta-obbt} is not desired, one
may simply set
$$\underline\beta_j = \underline\beta_j^0, \quad
\overline\beta_j = \overline\beta_j^0.$$

\subsubsection{Bounds on Signed Margins and Sigmoid Variables}

We now derive bounds on the
variables~$\theta$, $\alpha$, and~$h$ via
propagating the bounds on~$\beta$ obtained above.

First, for a clean observation, the signed
margin~$\theta_i^{tr} = s_i^{tr}\beta^\top x_i^{tr}$
is determined by~$\beta$. Since every
feasible~$\beta$ belongs to~$\mathcal{B}$, the
range of~$\theta_i^{tr}$ is contained in the interval
$[\underline\theta_i^{tr},\,\overline\theta_i^{tr}]$
given by
\begin{equation}
  \label{eq:theta-tr-bounds}
  \underline\theta_i^{tr}
  \define
  \min_{\beta\in\mathcal{B}}
  s_i^{tr}\beta^\top x_i^{tr},
  \quad
  \overline\theta_i^{tr}
  \define
  \max_{\beta\in\mathcal{B}}
  s_i^{tr}\beta^\top x_i^{tr}.
\end{equation}
For a poisoning point, the signed
margin~$\theta_k^p = s_k^p\beta^\top x_k^p$ depends on the decision variables~$\beta$ and~$x_k^p$.
Consider~$\xi^p = \beta^\top x^p$.
By the Cauchy--Schwarz inequality, $\|\beta\|_2 \leq R$ and~$\|x^p\|_2 \leq \sqrt{d}$, we have that $ |\xi^p|  \leq R\sqrt{d}.$

A tighter bound can be obtained componentwise.
Since we have $x_j^p \in [0,1]$ and~$\beta_j~\in~
[\underline\beta_j, \overline\beta_j]$, each
product~$\beta_j x_j^p$ satisfies
\[
  \beta_j x_j^p
  \in
  \bigl[\min\{0,\underline\beta_j\},\;
  \max\{0,\overline\beta_j\}\bigr].
\]
Summing over~$j$ and intersecting with the global
bound gives~$\xi^p \in [\underline \xi^p, \overline \xi^p]$
with
\begin{align}
  \underline \xi^p
  &\define
    \max\left\{
    -R\sqrt{d},\;
    \sum_{j=1}^d
    \min\{0,\underline\beta_j\}
    \right\},
    \label{eq:zp-lower}\\
  \overline \xi^p
  &\define
    \min\left\{
    R\sqrt{d},\;
    \sum_{j=1}^d
    \max\{0,\overline\beta_j\}
    \right\}.
    \label{eq:zp-upper}
\end{align}

The signed margin is~$\theta_k^p = s_k^p \xi^p$.
When~$s_k^p = 1$, the bounds on~$\xi^p$ apply directly.
When~$s_k^p = -1$, multiplying by~$-1$ reverses the
interval.  Both cases are captured by
\begin{equation}
  \label{eq:theta-p-bounds}
  \underline\theta_k^p
  \define
  \min\left\{
    s_k^p\underline \xi^p,\,
    s_k^p\overline \xi^p
  \right\},
  \quad
  \overline\theta_k^p
  \define
  \max\left\{
    s_k^p\underline \xi^p,\,
    s_k^p\overline \xi^p
  \right\}.
\end{equation}

If the tightest poisoning-margin bounds implied by~$\mathcal{B}$ are
desired, they can instead be obtained from
\begin{align}
  \underline \xi_{\mathcal{B}}^p
  &\define
    \min_{v\in\{1\}\times\{0,1\}^{d-1}}
    \;
    \min_{\beta\in\mathcal{B}}
    \beta^\top v,
    \label{eq:zp-exact-lower}\\
  \overline \xi_{\mathcal{B}}^p
  &\define
    \max_{v\in\{1\}\times\{0,1\}^{d-1}}
    \;
    \max_{\beta\in\mathcal{B}}
    \beta^\top v.
    \label{eq:zp-exact-upper}
\end{align}

Indeed, for fixed~$\beta$, a linear function of
$x\in\{1\}\times[0,1]^{d-1}$ attains its minimum
and maximum at a vertex of this admissible box.
Each inner problem in
\eqref{eq:zp-exact-lower}--\eqref{eq:zp-exact-upper} is convex.
The price for these exact bounds is the possible enumeration of
$2^{d-1}$ admissible vertices; hence~\eqref{eq:zp-lower}--\eqref{eq:zp-upper}
provide a substantially cheaper alternative.

Next, bounds on~$\theta$ propagate to bounds
on~$\alpha$.  Since~$\alpha_i = \phi'(\theta_i)$ at
every feasible point  and~$\phi'$ is strictly increasing, $\theta \in
[\underline\theta, \overline\theta]$ implies
$\alpha \in [\phi'(\underline\theta),
\phi'(\overline\theta)]$.
Explicitly,
\begin{align}
  \underline\alpha_i^{tr}
  &\define
    \phi'(\underline\theta_i^{tr})
    \leq
    \alpha_i^{tr}
    \leq
    \phi'(\overline\theta_i^{tr})
    \define
    \overline\alpha_i^{tr},
    \label{eq:alpha-tr-bounds}\\
  \underline\alpha_k^p
  &\define
    \phi'(\underline\theta_k^p)
    \leq
    \alpha_k^p
    \leq
    \phi'(\overline\theta_k^p)
    \define
    \overline\alpha_k^p.
    \label{eq:alpha-p-bounds}
\end{align}
Since~$\phi'(\theta) \in (0,1)$ for every
finite~$\theta$, in particular
\[
  0 \leq
  \underline\alpha_k^p
  \leq
  \alpha_k^p
  \leq
  \overline\alpha_k^p \leq  1.
\]

Finally, the bounds on~$h$ follow directly
from~\eqref{eq:h-box} and the upper bound
on~$\alpha_k^p$:
\begin{equation}
  \label{eq:h-bounds}
  0
  \leq
  h_{k,j}
  \leq
  \alpha_k^p
  \leq
  \overline\alpha_k^p.
\end{equation}

The bounds
\eqref{eq:beta-obbt},
\eqref{eq:theta-p-bounds},
\eqref{eq:alpha-p-bounds}, and
\eqref{eq:h-bounds}
provide finite intervals for every variable that
appears in a bilinear term. These bounds are derived
from the problem data and the regularization
parameter~$\lambda$. Thus, no arbitrary big-$M$ constants are needed.

\subsection{Equivalent Reformulated Problem}
\label{sec:final-reformulation}

Collecting the preceding results yields the alternative
single-level reformulation
\begin{varsubequations}{P3}
  \label{PA_3}
  \begin{alignat}{2}
    \max_{\beta,\theta^{tr},\theta^p, \alpha^{tr},
      \alpha^p,h}
    &
    -\sum_{i=1}^n \phi^*(\alpha_i^{tr})
    -\sum_{k=1}^q
    \bigl[\phi(\theta_k^p)+\phi^*(\alpha_k^p)\bigr]
    -2\lambda\|\beta\|_2^2
    &
    \label{eq:P3_obj}
    \\[4pt]
    \st\quad
    &
    \theta_i^{tr}
    = s_i^{tr}\beta^\top x_i^{tr},
    \quad
    i=1,\dots,n,
    \label{eq:P3_theta_tr}
    \\
    &
    0 \leq h_{k,j} \leq \alpha_k^p,     \quad
    k=1,\dots,q,\; j=2,\dots,d, h_{k,1} = \alpha_k^p,
    \label{eq:P3_hbox}
    \\
    &
    \alpha_k^p\,\theta_k^p
    = s_k^p\beta^\top h_k,
    \quad
    k=1,\dots,q,
    \label{eq:P3_bilinear}
    \\
    &
    \sum_{i=1}^n s_i^{tr}x_{i,j}^{tr}\alpha_i^{tr}
    +\sum_{k=1}^q s_k^p h_{k,j}
    +2\lambda\beta_j = 0,
    \quad
    j=1,\dots,d,
    \label{eq:P3_stationarity}
    \\
    &
    \sum_{i=1}^n
    \bigl[\phi(\theta_i^{tr})+\phi^*(\alpha_i^{tr})\bigr]
    +\sum_{k=1}^q
    \bigl[\phi(\theta_k^p)+\phi^*(\alpha_k^p)\bigr]
    +\,2\lambda\|\beta\|_2^2
    \leq 0,
    \label{eq:P3_gap}
    \\
    &
    \underline\beta_j
    \leq \beta_j
    \leq \overline\beta_j,
    \quad
    j=1,\dots,d,
    \label{eq:P3_beta_bounds}
    \\ &
    \underline\theta_i^{tr}
    \leq \theta_i^{tr}
    \leq \overline\theta_i^{tr},
    \quad
    i=1,\dots,n,
    \label{eq:P3_theta_tr_bounds}
    \\
    &
    \underline\theta_k^p
    \leq \theta_k^p
    \leq \overline\theta_k^p,
    \quad
    k=1,\dots,q,
    \label{eq:P3_theta_p_bounds}
    \\
    &
    \underline\alpha_i^{tr}
    \leq \alpha_i^{tr}
    \leq \overline\alpha_i^{tr},
    \quad
    i=1,\dots,n,
    \label{eq:P3_alpha_tr_bounds}
    \\
    &
    \underline\alpha_k^p
    \leq \alpha_k^p
    \leq \overline\alpha_k^p,
    \quad
    k=1,\dots,q.
    \label{eq:P3_alpha_p_bounds}
  \end{alignat}
\end{varsubequations}

\begin{prop}
  \label{prop:P2-P3-equivalence}
  Formulations~\eqref{PA_2} and~\eqref{PA_3} are equivalent.
  More precisely, every feasible point of~\eqref{PA_2} can be
  extended to a feasible point of~\eqref{PA_3} with the same
  objective value, and every feasible point of~\eqref{PA_3}
  determines feasible poisoning features for~\eqref{PA_2} with the
  same objective value.
\end{prop}

\begin{proof}
  First, let $(x^p,\beta)$ be feasible for~\eqref{PA_2}.
  Define
  \[
    \theta_i^{tr}
    =
    s_i^{tr}\beta^\top x_i^{tr},
    \quad
    \theta_k^p
    =
    s_k^p\beta^\top x_k^p,
    \quad
    \alpha_i^{tr}
    =
    \sigma(\theta_i^{tr}),
    \quad
    \alpha_k^p
    =
    \sigma(\theta_k^p),
    \quad
    h_k
    =
    \alpha_k^p x_k^p,
  \]
  with $h_{k,1} = \alpha_k^p x_{k,1}^p = \alpha_k^p$
  for every $k=1,\dots,q$.
  Then, \eqref{eq:P3_theta_tr}--\eqref{eq:P3_bilinear} hold.
  Moreover, $\alpha_i = \phi'(\theta_i)$ at every
  feasible point, which has two consequences.
  First, by~\eqref{eq:signed-loss},
  \[
    s_i^{tr}\alpha_i^{tr}
    =
    m_\beta(x_i^{tr})-y_i^{tr},
    \quad
    s_k^p\alpha_k^p
    =
    m_\beta(x_k^p)-y_k^p,
  \]
  so~\eqref{eq:P3_stationarity} is
  precisely~\eqref{eq:P2_kkt}.
  Second, Proposition~\ref{prop:logistic-conjugate}
  guarantees that every Fenchel gap~\eqref{eq:fenchel}
  vanishes, so~\eqref{eq:P3_gap} holds with equality.
  The bound
  constraints~\eqref{eq:P3_beta_bounds}--\eqref{eq:P3_alpha_p_bounds}
  are satisfied because~$\beta$ is a lower-level
  minimizer and the bounds in
  Section~\ref{subsec:valid-bounds} were derived for
  every such minimizer.
  Finally, the objective value of~\eqref{PA_3}
  equals~$\mathcal{L}(\mathcal{D}^{tr},\beta)$
  by~\eqref{eq:concave-objective}, which is the
  objective of~\eqref{PA_2}.

  Conversely, let
  $(\beta,\theta^{tr},\theta^p,
  \alpha^{tr},\alpha^p,h)$
  be feasible for~\eqref{PA_3}.
  Taking the scalar product of
  \eqref{eq:P3_stationarity} with~$\beta$ and using
  \eqref{eq:P3_theta_tr} and
  \eqref{eq:P3_bilinear} gives
  \[
    2\lambda\|\beta\|_2^2
    =
    -\sum_{i=1}^n
    \alpha_i^{tr}\theta_i^{tr}
    -
    \sum_{k=1}^q
    \alpha_k^p\theta_k^p.
  \]
  Substituting this into the left-hand side
  of~\eqref{eq:P3_gap} yields the sum of all
  individual Fenchel gaps:
  \begin{equation}\label{eq:ind_fenchel}
    \sum_{i=1}^n
    \bigl[
    \phi(\theta_i^{tr})
    +\phi^*(\alpha_i^{tr})
    -\alpha_i^{tr}\theta_i^{tr}
    \bigr]
    +
    \sum_{k=1}^q
    \bigl[
    \phi(\theta_k^p)
    +\phi^*(\alpha_k^p)
    -\alpha_k^p\theta_k^p
    \bigr]
    \leq 0.
  \end{equation}
  By  Proposition~\ref{prop:logistic-conjugate}, each
  Fenchel gap satisfies
  $\phi(\theta) + \phi^*(\alpha) - \alpha\theta \geq 0$  for
  any~$\alpha \in [0,1]$ and~$\theta \in \R$.
  Hence, every summand is nonnegative, which means the
  sum is also~$\geq 0$.  Combined with~$\leq 0$ in
  \eqref{eq:ind_fenchel}, the sum equals zero.  Since each summand is
  nonnegative and their sum is zero, each must vanish:
  \begin{align*}
    \phi(\theta_i^{tr})
    + \phi^*(\alpha_i^{tr})
    - \alpha_i^{tr}\theta_i^{tr} &= 0,
                                   \quad i = 1,\dots,n,
    \\
    \phi(\theta_k^{p})
    + \phi^*(\alpha_k^{p})
    - \alpha_k^{p}\theta_k^{p} &= 0,
                                 \quad k = 1,\dots,q.
  \end{align*}
  By Proposition~\ref{prop:logistic-conjugate},
  equality holds in~\eqref{eq:fenchel} if and only
  if~\eqref{eq:alpha-sigmoid} is satisfied.
  Since every gap vanishes, we conclude
  \begin{equation}\label{eq:alpha=}
    \alpha_i^{tr} = \phi'(\theta_i^{tr}),
    \quad
    \alpha_k^p = \phi'(\theta_k^p).
  \end{equation}
  In particular, $\alpha_k^p > 0$ holds, so we recover
  the poisoning features by setting
  \begin{equation}
    \label{eq:recover-xp}
    x_{k,j}^p \define \frac{h_{k,j}}{\alpha_k^p}.
  \end{equation}

  By~\eqref{eq:P3_hbox}, $x_k^p \in [0,1]^d$, with
    $x_{k,1}^p = 1$ for every $k=1,\dots,q$.

  Dividing~\eqref{eq:P3_bilinear} by~$\alpha_k^p$
  and using~\eqref{eq:recover-xp} gives
  \[
    \theta_k^p
    =
    s_k^p\,\beta^\top
    \frac{h_k}{\alpha_k^p}
    =
    s_k^p\,\beta^\top x_k^p.
  \]

  It remains to show that~$(x^p,\beta)$ is feasible
  for~\eqref{PA_2} and that the objective values
  coincide.
  As shown in the forward direction, using~\eqref{eq:alpha=},
  \eqref{eq:signed-loss} implies
  \[
    s_i^{tr}\alpha_i^{tr}
    = m_\beta(x_i^{tr}) - y_i^{tr},
    \quad
    s_k^p\alpha_k^p
    = m_\beta(x_k^p) - y_k^p.
  \]
  Substituting into~\eqref{eq:P3_stationarity}
  recovers~\eqref{eq:P2_kkt}, so~$(x^p,\beta)$ is
  feasible for~\eqref{PA_2}.

  Finally, since all Fenchel gaps vanish,
  \eqref{eq:P3_gap} holds with equality.
  By~\eqref{eq:concave-objective}, the objective
  of~\eqref{PA_3}
  equals~$\mathcal{L}(\mathcal{D}^{tr},\beta)$,
  which is the objective of~\eqref{PA_2}.
\end{proof}

\subsection{Convexity Structure and Spatial Branching}

The reformulation~\eqref{PA_3} isolates the nonconvexity of the
poisoning problem in constraint~\eqref{eq:P3_bilinear}.  More
precisely, we have the following:

\begin{itemize}
\item The maximization objective~\eqref{eq:P3_obj} is concave.

\item Constraints~\eqref{eq:P3_theta_tr},
  \eqref{eq:P3_hbox}, and
  \eqref{eq:P3_stationarity} are affine.

\item Constraint~\eqref{eq:P3_gap} is convex, since~$\phi$,
  $\phi^*$, and $\|\cdot\|_2^2$ are convex.

\item The only nonconvex constraint family is
  \eqref{eq:P3_bilinear}.  It contains the products
  $\alpha_k^p\theta_k^p$ and
  $\beta_jh_{k,j}$.
\end{itemize}

Thus, if~\eqref{eq:P3_bilinear}
were removed, the remaining problem would be convex.
In particular, neither~$\phi$ nor~$\phi^*$ requires
spatial branching and both can be represented using exponential-cone
constraints if supported by the solver.

For spatial branch-and-bound, it is convenient to lift the remaining
bilinearities. To this end, we introduce
\begin{equation}
  \label{eq:bilinear-lifts}
  r_k
  =
  \alpha_k^p\theta_k^p,
  \quad
  w_{k,j}
  =
  \beta_j h_{k,j}.
\end{equation}
Then~\eqref{eq:P3_bilinear} becomes the linear equation
\begin{equation}
  \label{eq:bilinear-link-linear}
  r_k
  =
  s_k^p
  \sum_{j=1}^d w_{k,j},
  \quad k=1,\dots,q,
\end{equation}
and the only nonconvexities are ones in~\eqref{eq:bilinear-lifts}.

For a generic bilinear constraint
\[
  w=uv,
  \quad
  u\in[\underline u,\overline u],
  \quad
  v\in[\underline v,\overline v],
\]
its convex hull over the bounding box is given by the McCormick
inequalities \parencite{mccormick1976computability}
\begin{align}
  w
  &\geq
    \underline u\,v
    +
    \underline v\,u
    -
    \underline u\,\underline v,
    \label{eq:mccormick1}\\
  w
  &\geq
    \overline u\,v
    +
    \overline v\,u
    -
    \overline u\,\overline v,
    \label{eq:mccormick2}\\
  w
  &\leq
    \overline u\,v
    +
    \underline v\,u
    -
    \overline u\,\underline v,
    \label{eq:mccormick3}\\
  w
  &\leq
    \underline u\,v
    +
    \overline v\,u
    -
    \underline u\,\overline v.
    \label{eq:mccormick4}
\end{align}
For the product $r_k=\alpha_k^p\theta_k^p$, the required
bounds are
\[
  \alpha_k^p
  \in
  [\underline\alpha_k^p,\overline\alpha_k^p],
  \quad
  \theta_k^p
  \in
  [\underline\theta_k^p,\overline\theta_k^p],
\]
and for the product $w_{k,j}=\beta_jh_{k,j}$, they are
\[
  \beta_j
  \in
  [\underline\beta_j,\overline\beta_j],
  \quad
  h_{k,j}
  \in
  [0,\overline\alpha_k^p],
\]
together with the stronger linear relation
$h_{k,j}\leq\alpha_k^p$.

If explicit bounds on the lifted variables are desired, they follow
directly and are given by
\begin{equation*}
  r_k
  \in
    \left[
    \min
    \left\{
    \underline\alpha_k^p\underline\theta_k^p,\,
    \underline\alpha_k^p\overline\theta_k^p,\,
    \overline\alpha_k^p\underline\theta_k^p,\,
    \overline\alpha_k^p\overline\theta_k^p
    \right\},
    \max
    \left\{
    \underline\alpha_k^p\underline\theta_k^p,\,
    \underline\alpha_k^p\overline\theta_k^p,\,
    \overline\alpha_k^p\underline\theta_k^p,\,
    \overline\alpha_k^p\overline\theta_k^p
    \right\}
    \right]
\end{equation*}
as well as
\[
  w_{k,j}
  \in
  \left[
    \min\{0,\underline\beta_j\overline\alpha_k^p\},
    \max\{0,\overline\beta_j\overline\alpha_k^p\}
  \right].
\]


\section{Numerical Results}

\label{sec:numerical_results}

In this section, we evaluate the poisoning attack on logistic
regression and assess the computational benefits of the
convex-analytic reformulation~\eqref{PA_3}.
Section~\ref{subsec:datasets} describes the benchmark datasets,
Section~\ref{subsec:setup} the computational setup,
Section~\ref{subsec:eval} the evaluation criteria, and
Section~\ref{subsec:results} the results.

\subsection{Datasets}
\label{subsec:datasets}
We evaluate the attack on eight binary classification datasets
drawn from the \textsf{RDatasets}
collection~\parencite{rdatasets_project}, namely from the
\codename{MASS} and \codename{ISLR} packages \parencite{ISLR}, and from the UCI
Machine Learning Repository~\parencite{uci_repository}.  The
datasets are chosen to cover application domains where data
poisoning poses a realistic threat: clinical decision support
(\codename{Biopsy}, breast cancer diagnosis), financial
prediction (\codename{Smarket}, S\&P~500 direction), counterfeit
detection (\codename{Banknote}), radar surveillance
(\codename{Ionosphere}), and spam filtering
(\codename{Spambase}~\parencite{spambase}).  Spam filters are a
classical target of poisoning, since they are retrained on data
that an adversary can partly control; \codename{Spambase} was
also used by~\textcite{mei2015using} to evaluate poisoning attacks
on logistic regression.  The remaining instances
(\codename{fgl}, forensic glass identification;
\codename{College}, university classification; \codename{OJ},
brand choice) complement the benchmark with varying sample sizes
and feature dimensions.  Table~\ref{tab:datasets} lists the
final instances.

We apply the following preprocessing to every dataset.
First, observations with at least one missing value are
removed.  Second, the target variable is encoded as a
binary label $y \in \{0,1\}$; for \codename{fgl}, whose
original target distinguishes several glass types, we
merge the window-glass categories (\textsf{WinF}, \textsf{WinNF}) into the
positive class ($y=1$) and the remaining types into the
negative class ($y=0$). Third, since the attack model requires
features in a continuous box~$[0,1]^d$, only numeric features are
retained.  Columns with zero variance are removed
because they carry no discriminative information and
would produce a degenerate scaling.
For \codename{Smarket}, we use the predictors of the standard logistic
regression example of \parencite{ISLR} (\textsf{Lag1}--\textsf{Lag5},
\textsf{Volume}); in particular, \textsf{Today} is excluded, since the
label \textsf{Direction} is its sign.
For \codename{OJ}, and only there, numeric columns with at most ten
distinct values are removed: store identifiers and promotion
indicators encode categories, for which intermediate values have no
meaning, and list prices are linear combinations of retained features.
Ordinal scores, such as the predictors of \codename{biopsy} (integers
from $1$ to $10$), are retained and treated as continuous. Poisoning
points may thus take values that are not physically realizable; since
this relaxation only enlarges the attacker's feasible set, the
optimal training-loss objective value is an upper bound on that
attainable by realizable attacks.

All features are scaled to~$[0,1]$.  For each
coordinate~$j$, the minimum~$l_j = \min_i x_j^i$ and
maximum~$u_j = \max_i x_j^i$ are computed on the
training partition only, and the transformation
\[
  \tilde{x}_j^i = (x_j^i - l_j)/(u_j - l_j)
\]  is
applied. If $u_j = l_j$, the denominator is set to~$1$.
Test points are transformed with the same
parameters~$l_j$, $u_j$ and clipped to~$[0,1]$, so
that all points lie in the same box as the admissible
poisoning features.
After scaling, a constant column of ones is prepended
to encode the intercept, so that $x_{i,1} = 1$ for all
training, test, and poisoning points, as required by
the model in Section~\ref{sec:model}.

\begin{table}[t]
  \centering
  \caption{Benchmark datasets after preprocessing.  $m$~is the
    total number of observations, $d$~the number of numeric
    features (including the intercept column), and
    $P(y{=}1)$~the proportion of positive labels.}
  \label{tab:datasets}
  \begin{tabular}{llrrrl}
    \toprule
    Instance & Source & $m$ & $d$ & $P(y{=}1)$ & Domain \\
    \midrule
    \textsf{Biopsy}      & MASS & 683  & 10  & 0.35 & Healthcare (breast cancer) \\
    \textsf{Smarket}     & ISLR & 1250 & 7 & 0.52 & Finance (S\&P~500)         \\
    \textsf{fgl}         & MASS & 214  & 10  & 0.68 & Forensics (glass)          \\
    \textsf{College}     & ISLR & 777  & 18 & 0.73 & Education                  \\
    \textsf{OJ}          & ISLR & 1070 & 11 & 0.61 & Retail (brand choice)      \\
    \textsf{Banknote}    & UCI  & 1372 & 5  & 0.44 & Counterfeit detection      \\
    \textsf{Ionosphere}  & UCI  & 351  & 34 & 0.64 & Radar (surveillance)       \\
    \textsf{Spambase}    & UCI  & 4601 & 58 & 0.39 & Spam filtering            \\
    \bottomrule
  \end{tabular}
\end{table}

\subsection{Computational Setup}
\label{subsec:setup}

For each dataset, we perform a stratified split that assigns
30\,\% of the observations of each class to the training set and
the remaining observations to the test set, so that the
training set size is
\[
  n = \sum_{c \in \{0,1\}} \nint{0.3\, n_c},
\]
where $n_c$ denotes the number of observations with label~$c$
and $\nint{\cdot}$ denotes rounding to the nearest integer,
with ties rounded to the nearest even integer.
We consider poisoning budgets
$q = \max\{1, \nint{\rho\, n}\}$ with
$\rho \in \{5\,\%, 10\,\%, 15\,\%, 20\,\%\}$.
The labels of the
poisoned points are fixed in advance and alternate between the
two classes, so that $\lceil q/2 \rceil$ points have $y = 0$
and $\lfloor q/2 \rfloor$ have $y = 1$; the attacker only
chooses their features.  The regularization parameter is set
to $\lambda = 1$ throughout.

The experiments are organized into two phases, each addressing
a different question.

\begin{description}
\item[Phase~1: Attack Effectiveness]
  We first ask whether inserting poisoned points substantially
  degrades the classifier.  Answering this requires an effective
  attack rather than an optimal one, so a local solver suffices,
  which allows us to repeat the experiment over 5~data
  partitions.  For each partition, we compare:
  \begin{itemize}
  \item \textsf{LR}: the regularized logistic regression~\eqref{LR},
    solved with \textsf{Ipopt}~\parencite{Ipopt}.  Since~\eqref{LR}
    is strongly convex, a single run yields its unique global
    solution.
  \item \textsf{PA-L}: the KKT reformulation~\eqref{PA_2} with the
    bounds of Proposition~\ref{prop:beta-norm-bound}, solved with
    \textsf{Ipopt} from 5~random starting points, with
    $\beta \sim \mathcal{N}(0, 0.1^2 I)$ and the poisoned features
    drawn uniformly in~$[0,1]$.  Each run is limited to 300\,s,
    and the solution with the largest upper-level objective value
    is retained.
  \end{itemize}

\item[Phase~2: Global Optimization]
  The second question is how much the convex-analytic
  reformulation helps global optimization.  To answer it, we
  use one partition per dataset  and compare, with a time
  limit of 3600\,s per instance:
  \begin{itemize}
  \item \textsf{PA-G}: \textsf{SCIP} applied to the KKT
    reformulation~\eqref{PA_2}, with only the bounds of
    Proposition~\ref{prop:beta-norm-bound}.
  \item \textsf{CAR}: \textsf{SCIP} applied to the convex-analytic
    reformulation~\eqref{PA_3}, with only the bounds of
    Proposition~\ref{prop:beta-norm-bound}.
  \item \textsf{PA-G+}: \textsf{SCIP} applied to~\eqref{PA_2}, with all
    bounds of Section~\ref{subsec:valid-bounds}.
  \item \textsf{CAR+}: \textsf{SCIP} applied to~\eqref{PA_3}, with all
    bounds of Section~\ref{subsec:valid-bounds}.
  \end{itemize}
\end{description}
The convex problems~\eqref{eq:beta-obbt}
and~\eqref{eq:theta-tr-bounds} that provide the bounds of \textsf{PA-G+}
and \textsf{CAR+} are solved with \textsf{Ipopt} (tolerance $10^{-9}$), and
each computed bound~$v$ is relaxed by $10^{-6}(1+|v|)$. These bounds
are therefore numerical: they are valid up to the accuracy of
\textsf{Ipopt}, but not certified. The bounds of
Proposition~\ref{prop:beta-norm-bound}, used by \textsf{PA-G} and
\textsf{CAR}, are computed by scalar bisection.

Finally, to avoid evaluating $\log 0$ in~$\phi^*$, the bounds
on~$\alpha$ in \textsf{CAR} and \textsf{CAR+} are clipped
to~$[10^{-8}, 1-10^{-8}]$.
All models are implemented in \textsf{Julia}~1.12.3
\parencite{bezanson2017julia} using
\textsf{JuMP}~1.29.3~\parencite{DunningHuchetteLubin2017}, with
\textsf{SCIP}~10.0.3~\parencite{scip10} (relative gap $10^{-4}$, feasibility
tolerance $10^{-6}$) and \textsf{Ipopt}~3.14.19 (tolerance
$10^{-6}$ for \textsf{PA-L}). Six runs are executed in parallel, with one
thread each, on an Apple M3 Pro (12 cores: 6 performance and
6 efficiency) with 36\,GB unified memory. Phase~1 uses five random
partitions (seeds $42$--$46$); Phase~2 uses the first of them
(seed~$42$), and the five starting points of \textsf{PA-L} are drawn with
seeds $1000p+s$, where $p$ is the partition seed
and $s=1,\ldots,5$ indexes the starts.
The code is publicly available at
\url{https://github.com/mariaepinheiro/Poisoning_Attacks_Logistic_Regression}.

In preliminary experiments on a subset of the instances,
\textsf{Gurobi}~13.0~\parencite{gurobi} obtained
substantially weaker dual bounds than \textsf{SCIP} for both
formulations.  We therefore report results only for
\textsf{SCIP}, which has the additional advantage of being
open source, so that all results can be reproduced without a
commercial license.

\subsection{Evaluation Criteria}
\label{subsec:eval}

\subsubsection{Attack Effectiveness.}

A test point~$x$ is assigned to the positive class whenever
$m_\beta(x) \geq 0.5$ and to the negative one otherwise.
Knowing the true labels, we categorize predictions as true
positive~(TP), true negative~(TN), false positive~(FP), or
false negative~(FN), and compute two metrics for which a higher
value indicates a better classification.
From the resulting true positives~(TP), true negatives~(TN),
false positives~(FP), and false negatives~(FN), we compute the
accuracy
\begin{equation}
  \label{eq:AC}
  \AC \define \frac{\TP + \TN}{\TP + \TN + \FP + \FN}
  \in [0,1],
\end{equation}
and the Matthews correlation
coefficient~\parencite{chicco2023matthews}
\begin{equation}
  \label{eq:MCC}
  \MCC \define
  \frac{\TP \cdot \TN - \FP \cdot \FN}
  {\sqrt{(\TP+\FP)(\TP+\FN)(\TN+\FP)(\TN+\FN)}}
  \in [-1,1],
\end{equation}
with $\MCC = 0$ when the denominator vanishes.  For both
metrics, higher values indicate better classification.
Let $\AC_{\text{\textsf{LR}}}$ and $\MCC_{\text{\textsf{LR}}}$ denote the test
metrics of the clean model \textsf{LR}, and $\AC_{\text{\textsf{PA-L}}}$ and
$\MCC_{\text{\textsf{PA-L}}}$ those of the poisoned model returned by
\textsf{PA-L}.  The effect of the attack is measured by the drops
\begin{equation}\label{eq:drop}
  \Delta\AC = \AC_{\text{\textsf{LR}}} - \AC_{\text{\textsf{PA-L}}},
  \quad
  \Delta\MCC = \MCC_{\text{\textsf{LR}}} - \MCC_{\text{\textsf{PA-L}}},
\end{equation}
computed for each partition; a positive drop indicates that
the attack degraded the classifier.  To assess the
computational cost of the attack, we also record the
wall-clock times of \textsf{LR} and \textsf{PA-L} in seconds.  All quantities
are reported as mean~$\pm$~standard deviation over the
5~partitions.

\subsubsection{Global Optimization.}

By Proposition~\ref{prop:P2-P3-equivalence}, Problems~\eqref{PA_2}
and~\eqref{PA_3} are equivalent maximization problems with optimal
value~$z^*$.  The bounds of Proposition~\ref{prop:beta-norm-bound} do not exclude
any feasible point, and those of Section~\ref{subsec:valid-bounds} are
valid up to the accuracy of \textsf{Ipopt}. For \textsf{CAR} and \textsf{CAR+}, the
clipping of~$\alpha$ to $[10^{-8}, 1-10^{-8}]$ is needed because
\textsf{SCIP} cannot evaluate $\log$ at values below its tolerance.
However, it restricts the feasible set, so their dual bounds refer to this
restricted problem.

To assess the bounds on the variables, we report, for each set of
bounds, the average width of the intervals obtained for~$\beta_j$,
$\theta_i^{tr}$, $\xi^p$, and~$\alpha_k^p$, where the average is taken
over all components and over the four budgets~$\rho$.  We also report
the time needed to compute all bounds of
Section~\ref{subsec:valid-bounds}, which is spent before \textsf{SCIP}
is started and is not included in its time limit.

For each method~$f$, \textsf{SCIP} reports a \emph{primal
  bound}~$\underline{z}_f$, the objective value of the best feasible
solution found, and a \emph{dual bound}~$\overline{z}_f$, obtained
from relaxations, so that
\begin{equation}
  \label{eq:bounds_gap}
  \underline{z}_f \leq z^* \leq \overline{z}_f.
\end{equation}

In addition, \eqref{eq:clean-valid-ineq} yields the analytic bound
$z^* \le \bar U := N\log 2 - q\,\phi(-R\sqrt{d})$, which requires no
optimization. Since $\min\{\overline{z}_f, \bar U\}$ is also an upper
bound on~$z^*$, we use it in place of~$\overline{z}_f$ below.

As a method may fail to find a feasible solution within the time
limit, we use the best primal bound of all four methods,
\[
  z_{\mathrm{best}} =
  \max\{\underline{z}_{\text{\textsf{PA-G}}},\,
  \underline{z}_{\text{\textsf{CAR}}},\,
  \underline{z}_{\text{\textsf{PA-G+}}},\,
  \underline{z}_{\text{\textsf{CAR+}}}\},
\]
as a common reference, and measure the distance of each dual bound
from it by
\begin{equation}
  \label{eq:gap}
  \mathrm{gap}_f =
  \frac{\min\{\overline{z}_f, \bar U\} - z_{\mathrm{best}}}{|z_{\mathrm{best}}|}.
\end{equation}

By~\eqref{eq:bounds_gap}, a gap of~$0$ certifies global optimality,
and a gap of, say, $0.2$ guarantees that $z^*$ exceeds the best attack
found by at most~$20\,\%$.  Since all methods share the same reference,
their gaps are directly comparable.

To compare the dual bounds obtained for~\eqref{PA_2} and~\eqref{PA_3}
independently of the primal bounds, we also report, for each set of
bounds, the ratio
\begin{equation}
  \label{eq:ratio}
  r = \frac{\min\{\overline{z}_{\text{\textsf{PA-G}}}, \bar U\}}
             {\min\{\overline{z}_{\text{\textsf{CAR}}}, \bar U\}},
  \quad
  r_+ = \frac{\min\{\overline{z}_{\text{\textsf{PA-G+}}}, \bar U\}}
           {\min\{\overline{z}_{\text{\textsf{CAR+}}}, \bar U\}}.
\end{equation}
A ratio greater than~$1$ means that the corresponding convex-analytic
method provides the tighter upper bound on~$z^*$ within the time
limit. For each method, we report
$\underline{z}_f$, $\overline{z}_f$, and $\mathrm{gap}_f$ in percent,
together with the analytic bound~$\bar U$ and the ratios~$r$ and~$r_+$.


\subsection{Discussion of the Results}
\label{subsec:results}

\subsubsection{Attack Effectiveness}
\label{subsec:effectiveness}

\begin{figure}[t]
  \centering
  \includegraphics[width=\textwidth]{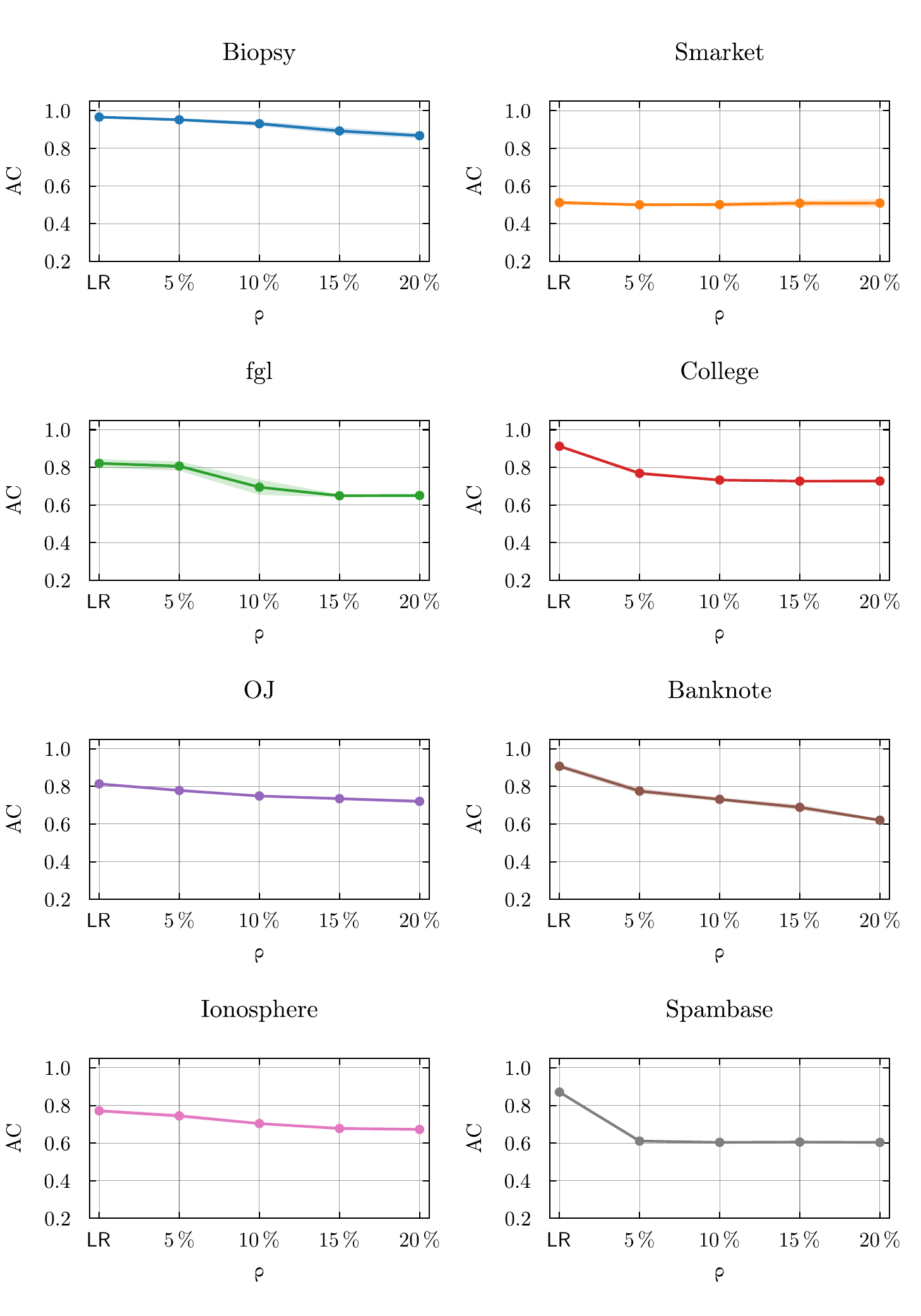}
  \caption{Test accuracy of the clean model (\textsf{LR}) and of the
    poisoned models returned by \textsf{PA-L} for each budget~$\rho$.
    Markers: mean over 5~partitions; shaded areas:
    $\pm$~one standard deviation.}
  \label{fig:ac_grid}
\end{figure}

\begin{figure}[t]
  \centering
  \includegraphics[width=\textwidth]{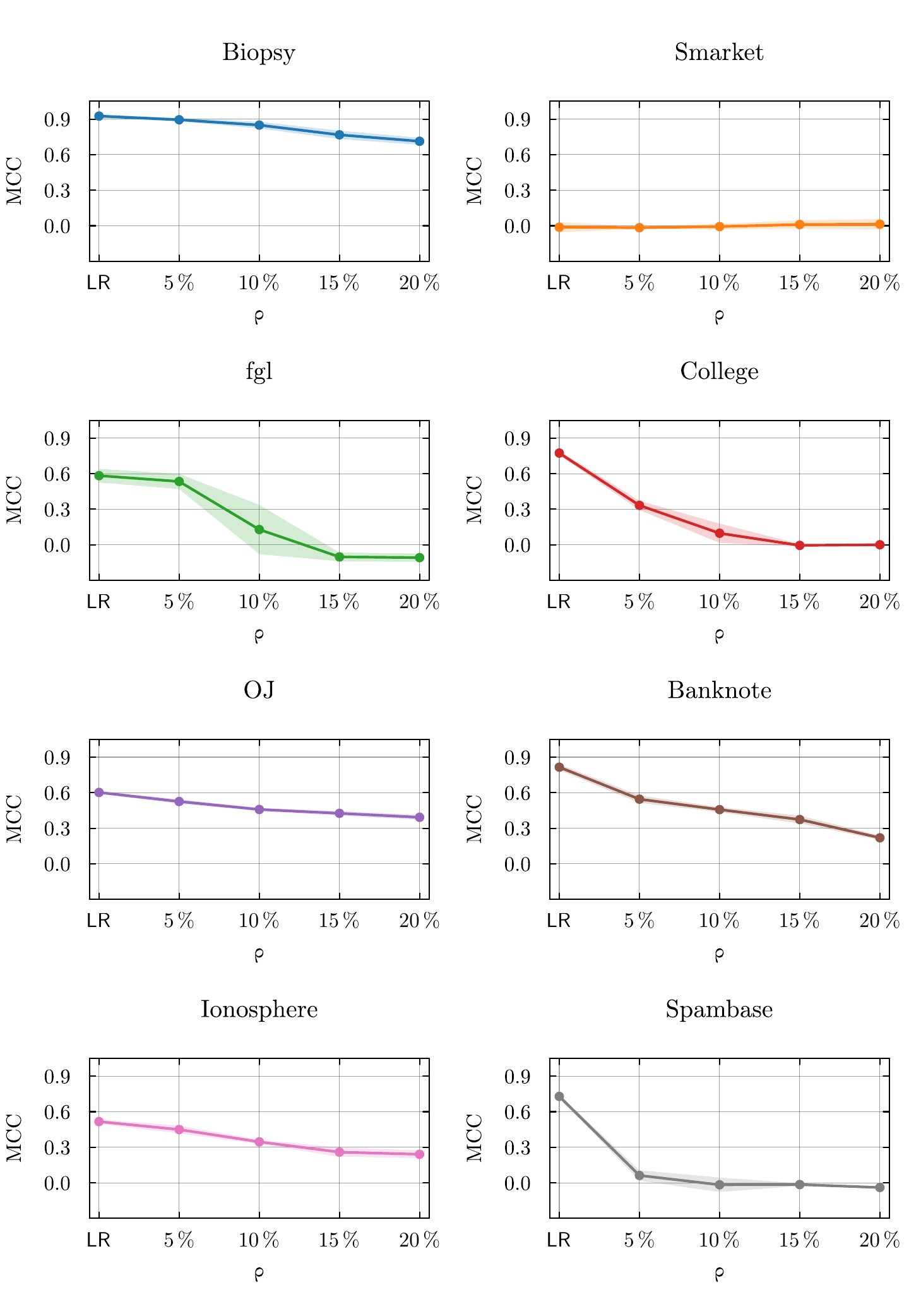}
  \caption{Test MCC of the clean model (\textsf{LR}) and of the
    poisoned models returned by \textsf{PA-L} for each budget~$\rho$.
    Markers: mean over 5~partitions; shaded areas:
    $\pm$~one standard deviation.}
  \label{fig:mcc_grid}
\end{figure}

Figures~\ref{fig:ac_grid} and~\ref{fig:mcc_grid} show the test
accuracy and MCC of the \textsf{LR} model and of the poisoned
models returned by \textsf{PA-L} for each budget~$\rho$.
Table~\ref{tab:attack_drops} reports the corresponding drops
and solve times.

\begin{table}[t]
\centering
\footnotesize
\caption{Attack effectiveness: drops $\Delta\AC$ and $\Delta\MCC$ as in \eqref{eq:drop} induced by
  \textsf{PA-L}, computed per partition, and solve times in seconds; \textsf{PA-L} times include
  all 5~starts.  All values: mean~$\pm$~standard deviation over 5~partitions.}
\label{tab:attack_drops}
\begin{tabular}{@{}l r c rrr@{}}
\toprule
& \multicolumn{1}{c}{\textsf{LR}} & & \multicolumn{3}{c}{\textsf{PA-L}} \\
\cmidrule(lr){2-2} \cmidrule(lr){4-6}
Dataset & Time (s) & $\rho$ & $\Delta\AC$ & $\Delta\MCC$ & Time (s) \\
\midrule
\multirow{4}{*}{\textsf{Biopsy}} & \multirow{4}{*}{$0.003 \pm 0.001$} & 5\,\% & $0.014 \pm 0.003$ & $0.031 \pm 0.008$ & $0.179 \pm 0.027$ \\
 &  & 10\,\% & $0.035 \pm 0.010$ & $0.076 \pm 0.021$ & $0.183 \pm 0.022$ \\
 &  & 15\,\% & $0.074 \pm 0.015$ & $0.158 \pm 0.031$ & $0.165 \pm 0.039$ \\
 &  & 20\,\% & $0.099 \pm 0.013$ & $0.212 \pm 0.028$ & $0.152 \pm 0.019$ \\
\midrule
\multirow{4}{*}{\textsf{Smarket}} & \multirow{4}{*}{$0.003 \pm 0.001$} & 5\,\% & $0.012 \pm 0.011$ & $0.004 \pm 0.039$ & $0.152 \pm 0.011$ \\
 &  & 10\,\% & $0.011 \pm 0.020$ & $-0.005 \pm 0.059$ & $0.166 \pm 0.021$ \\
 &  & 15\,\% & $0.003 \pm 0.025$ & $-0.022 \pm 0.073$ & $0.197 \pm 0.035$ \\
 &  & 20\,\% & $0.003 \pm 0.028$ & $-0.025 \pm 0.080$ & $0.201 \pm 0.022$ \\
\midrule
\multirow{4}{*}{\textsf{fgl}} & \multirow{4}{*}{$0.002 \pm 0.001$} & 5\,\% & $0.015 \pm 0.019$ & $0.049 \pm 0.050$ & $0.039 \pm 0.007$ \\
 &  & 10\,\% & $0.127 \pm 0.035$ & $0.455 \pm 0.188$ & $0.042 \pm 0.007$ \\
 &  & 15\,\% & $0.172 \pm 0.029$ & $0.685 \pm 0.085$ & $0.046 \pm 0.004$ \\
 &  & 20\,\% & $0.171 \pm 0.027$ & $0.692 \pm 0.083$ & $0.056 \pm 0.006$ \\
\midrule
\multirow{4}{*}{\textsf{College}} & \multirow{4}{*}{$0.011 \pm 0.002$} & 5\,\% & $0.144 \pm 0.014$ & $0.441 \pm 0.049$ & $0.575 \pm 0.062$ \\
 &  & 10\,\% & $0.180 \pm 0.010$ & $0.677 \pm 0.090$ & $0.662 \pm 0.072$ \\
 &  & 15\,\% & $0.185 \pm 0.006$ & $0.779 \pm 0.019$ & $0.712 \pm 0.113$ \\
 &  & 20\,\% & $0.185 \pm 0.007$ & $0.774 \pm 0.018$ & $0.732 \pm 0.081$ \\
\midrule
\multirow{4}{*}{\textsf{OJ}} & \multirow{4}{*}{$0.006 \pm 0.002$} & 5\,\% & $0.035 \pm 0.012$ & $0.077 \pm 0.028$ & $0.293 \pm 0.022$ \\
 &  & 10\,\% & $0.064 \pm 0.009$ & $0.144 \pm 0.021$ & $0.347 \pm 0.027$ \\
 &  & 15\,\% & $0.079 \pm 0.009$ & $0.176 \pm 0.019$ & $0.351 \pm 0.020$ \\
 &  & 20\,\% & $0.093 \pm 0.010$ & $0.210 \pm 0.024$ & $0.367 \pm 0.028$ \\
\midrule
\multirow{4}{*}{\textsf{Banknote}} & \multirow{4}{*}{$0.003 \pm 0.001$} & 5\,\% & $0.132 \pm 0.007$ & $0.270 \pm 0.015$ & $0.207 \pm 0.014$ \\
 &  & 10\,\% & $0.176 \pm 0.007$ & $0.358 \pm 0.015$ & $0.165 \pm 0.011$ \\
 &  & 15\,\% & $0.218 \pm 0.006$ & $0.442 \pm 0.016$ & $0.137 \pm 0.006$ \\
 &  & 20\,\% & $0.287 \pm 0.011$ & $0.596 \pm 0.022$ & $0.129 \pm 0.009$ \\
\midrule
\multirow{4}{*}{\textsf{Ionosphere}} & \multirow{4}{*}{$0.011 \pm 0.000$} & 5\,\% & $0.027 \pm 0.009$ & $0.067 \pm 0.026$ & $0.200 \pm 0.014$ \\
 &  & 10\,\% & $0.068 \pm 0.013$ & $0.171 \pm 0.033$ & $0.235 \pm 0.028$ \\
 &  & 15\,\% & $0.094 \pm 0.017$ & $0.258 \pm 0.055$ & $0.293 \pm 0.026$ \\
 &  & 20\,\% & $0.099 \pm 0.014$ & $0.276 \pm 0.045$ & $0.337 \pm 0.024$ \\
\midrule
\multirow{4}{*}{\textsf{Spambase}} & \multirow{4}{*}{$0.095 \pm 0.012$} & 5\,\% & $0.260 \pm 0.012$ & $0.667 \pm 0.052$ & $68.059 \pm 15.025$ \\
 &  & 10\,\% & $0.267 \pm 0.010$ & $0.746 \pm 0.069$ & $8.226 \pm 0.145$ \\
 &  & 15\,\% & $0.265 \pm 0.006$ & $0.744 \pm 0.010$ & $8.520 \pm 0.318$ \\
 &  & 20\,\% & $0.267 \pm 0.007$ & $0.769 \pm 0.018$ & $10.072 \pm 0.225$ \\
\bottomrule
\end{tabular}
\end{table}

The attack degrades the mean accuracy in all datasets, although
the mean drops are small for \codename{Smarket}. The degradation
generally grows with the budget~$\rho$, reaching drops of up to $0.29$
(\codename{Banknote}). For \codename{Smarket}, the clean
model has mean accuracy $\AC = 0.512$, and the observed mean accuracy
drops range from $0.003$ to $0.012$.
As with accuracy, MCC generally decreases with the budget~$\rho$ in all
datasets except \codename{Smarket}, but its drops are larger and
reveal effects that accuracy hides.  For \codename{College}
($\rho \geq 15\,\%$) and \codename{Spambase} ($\rho \geq 10\,\%$), the
mean MCC of the poisoned model is close to zero.

The response to increasing~$\rho$ differs substantially between
datasets. At $\rho=5\,\%$, the mean accuracy drop is only $0.014$ for
\codename{Biopsy}, compared with $0.132$ for \codename{Banknote} and
$0.260$ for \codename{Spambase}. For \codename{Banknote}, the drop
continues to grow across the four budgets, whereas most of the
accuracy degradation on \codename{Spambase} already occurs at
$\rho=5\,\%$. The low accuracy on \codename{Smarket}, however, is already
present in the clean model and does not demonstrate greater
susceptibility to poisoning.

A plateau in test accuracy does not imply a plateau in the attacker's
objective. For \codename{Spambase}, the mean objective value returned
by \textsf{PA-L} increases from approximately $908$ at $\rho=5\,\%$ to $998$ at
$\rho=20\,\%$. The attacker maximizes logistic loss on the clean training
data, whereas accuracy and MCC evaluate thresholded predictions on
the test data. Predicted probabilities can change without changing
the predicted classes, and the training and test losses need not
respond in the same way.

Possible explanations for the differences between datasets include
the magnitude of the clean decision scores and the sensitivity of the
fitted coefficients to poisoning. Points that are correctly
classified with larger absolute decision scores (i.e., more
negative $\theta$ under the convention of
Section~\ref{sec:signed-margins}) require larger changes before
their predicted labels change. The coefficient sensitivity depends on the curvature of the lower-level objective
and the gradient contributions of admissible poisoning points; this
relationship underlies influence-function analysis
\parencite{koh2017understanding}. Moreover, the box constraint permits
feature combinations beyond those observed in the clean data, which
may give poisoning points substantial influence. These mechanisms
provide hypotheses for further investigation. The present experiments
do not measure their contributions and therefore do not establish a
general rule relating dataset properties to attack susceptibility.
Such comparisons also depend on scaling and regularization: with the
summed loss used here, a fixed~$\lambda$ corresponds to a penalty
weight~$\lambda/(n+q)$ relative to the average loss on the contaminated
training set.

The attack is also cheap to compute.  \textsf{LR} takes at most $0.1$\,s,
and \textsf{PA-L}, including all 5~starts, at most $0.73$\,s on average,
except for \codename{Spambase}, where it takes between $8$ and
$10$\,s, and about $68$\,s for $\rho = 5\,\%$.  We stress, however,
that these results come from local solutions of~\eqref{PA_2}, with
no guarantee of global optimality.

\subsubsection{Global Optimization}
\label{subsec:obj-gap}

Table~\ref{tab:bound_widths} reports the average width of the bounds
obtained from Proposition~\ref{prop:beta-norm-bound} alone and from all
bounds of Section~\ref{subsec:valid-bounds}. Computing all bounds takes
between $0.34$ and $434.88$\,s on average over the four budgets (last column of
Table~\ref{tab:bound_widths}); except for \codename{Spambase}, it stays
below $8$\,s. This time comes on top of the time limit of \textsf{SCIP}.

The additional bounds substantially tighten the quantities associated
with the training data: across the 32 individual
combinations of datasets and poisoning budgets, the width of the
bounds on~$\beta$ is reduced to between $18\,\%$ and $67\,\%$, and that of
the bounds on the margins~$\theta^{tr}$ to between $5\,\%$ and $38\,\%$ of
the width obtained from Proposition~\ref{prop:beta-norm-bound}.  In
contrast, the quantities associated with the poisoning points are
hardly affected. The bounds on~$\xi^p$, and hence on~$\theta^p$, are
unchanged in 24 of the 32~instances and only moderately tighter in the
others.  Since the sigmoid saturates quickly, the bounds
on~$\alpha^p$, and therefore on~$h$, remain essentially $[0,1]$ in all
instances.
\begin{table}[t]
\centering
\footnotesize
\caption{Average bound widths with the bounds of
  Proposition~\ref{prop:beta-norm-bound} only (Prop.~3) and with all bounds
  of Section~\ref{subsec:valid-bounds} (All), and time to compute all
  bounds.  Mean over the four budgets~$\rho$.}
\label{tab:bound_widths}
\begin{tabular}{@{}l rr rr rr rr r@{}}
\toprule
 & \multicolumn{2}{c}{$\beta$} & \multicolumn{2}{c}{$\theta^{tr}$}
 & \multicolumn{2}{c}{$\xi^p$} & \multicolumn{2}{c}{$\alpha^p$} & Time \\
\cmidrule(lr){2-3} \cmidrule(lr){4-5} \cmidrule(lr){6-7} \cmidrule(lr){8-9}
Dataset & Prop.~3 & All & Prop.~3 & All & Prop.~3 & All & Prop.~3 & All & (s) \\
\midrule
\textsf{Biopsy} & 17.87 & 11.69 & 26.03 & 9.69 & 56.51 & 56.51 & 1.000 & 1.000 & 1.46 \\
\textsf{Smarket} & 24.18 & 6.33 & 38.22 & 2.56 & 63.97 & 44.29 & 1.000 & 1.000 & 4.80 \\
\textsf{fgl} & 9.98 & 4.64 & 15.76 & 3.75 & 31.57 & 31.57 & 1.000 & 1.000 & 0.34 \\
\textsf{College} & 19.10 & 11.75 & 37.88 & 7.92 & 81.02 & 81.02 & 1.000 & 1.000 & 7.34 \\
\textsf{OJ} & 22.36 & 10.63 & 42.49 & 7.23 & 74.17 & 74.17 & 1.000 & 1.000 & 5.39 \\
\textsf{Banknote} & 25.34 & 10.03 & 38.74 & 4.69 & 56.66 & 49.88 & 1.000 & 1.000 & 3.63 \\
\textsf{Ionosphere} & 12.85 & 6.77 & 51.90 & 8.18 & 74.95 & 74.95 & 1.000 & 1.000 & 4.59 \\
\textsf{Spambase} & 46.38 & 25.41 & 53.37 & 11.80 & 353.20 & 353.20 & 1.000 & 1.000 & 434.88 \\
\bottomrule
\end{tabular}
\end{table}

Table~\ref{tab:obj_gap} reports the primal bounds, dual bounds, and
gaps~\eqref{eq:gap} of the four methods. \textsf{PA-G+} finds a feasible
solution in every instance, and its primal bound is within
approximately $0.02$ of
the maximum $z_{\mathrm{best}}$ over the four methods, which we use as
a common reference; it therefore does not depend on the solutions
found by \textsf{CAR+}. None of the 128~runs certifies global optimality: all
of them stop at the time limit of 3600\,s.

\begin{table}[p]
\centering
\caption{Primal bounds $\underline{z}_f$, dual bounds $\overline{z}_f$
  (\textsf{SCIP}, time limit 3600\,s), analytic bound~$\bar U$, and
  gaps~\eqref{eq:gap} in \,\%. Smallest gap of each instance in bold;
  ``--'' = no feasible solution found.}
\label{tab:obj_gap}
\scriptsize
\setlength{\tabcolsep}{3pt}
\begin{adjustbox}{max width=\textwidth}
\begin{tabular}{@{}ll rrrrrrrr@{}}
\toprule
 & & \textsf{Biopsy} & \textsf{Smarket} & \textsf{fgl} & \textsf{College} & \textsf{OJ} & \textsf{Banknote} & \textsf{Ionosphere} & \textsf{Spambase} \\
\midrule
\multicolumn{10}{@{}l}{$\rho = 5\,\%$} \\
\cmidrule(r){1-2}
\multirow{4}{*}{$\underline{z}_f$} & \textsf{PA-G} & 45.55 & 261.98 & 33.37 & 102.44 & 150.43 & 209.98 & 52.84 & 912.97 \\
 & \textsf{CAR} & 43.83 & 260.37 & 33.37 & 102.45 & 150.44 & 207.11 & 52.85 & -- \\
 & \textsf{PA-G+} & 45.55 & 261.98 & 33.37 & 102.44 & 150.43 & 209.98 & 52.84 & 912.97 \\
 & \textsf{CAR+} & 45.55 & 261.98 & 33.37 & 102.45 & 150.44 & 210.00 & 52.85 & -- \\
\addlinespace[2pt]
\multirow{4}{*}{$\overline{z}_f$} & \textsf{PA-G} & 3016.9 & 9190.6 & 475.4 & 6411.4 & 10223.3 & 9856.9 & 3843.3 & 48918.4 \\
 & \textsf{CAR} & 73.8 & 272.9 & 36.5 & 126.1 & 180.4 & 248.6 & 62.5 & 967.4 \\
 & \textsf{PA-G+} & 661.5 & 330.5 & 63.7 & 867.1 & 967.9 & 607.3 & 373.2 & 6728.7 \\
 & \textsf{CAR+} & 73.8 & 272.9 & 36.4 & 126.1 & 180.4 & 248.8 & 62.5 & 967.4 \\
\addlinespace[2pt]
 & $\bar U$ & 149.0 & 273.1 & 46.4 & 170.5 & 233.6 & 300.1 & 76.9 & 1004.4 \\
\addlinespace[2pt]
\multirow{4}{*}{gap (\,\%)} & \textsf{PA-G} & 227.1 & \textbf{4.2} & 39.2 & 66.4 & 55.3 & 42.9 & 45.6 & 10.0 \\
 & \textsf{CAR} & 62.1 & \textbf{4.2} & 9.5 & \textbf{23.1} & \textbf{19.9} & \textbf{18.4} & \textbf{18.3} & \textbf{6.0} \\
 & \textsf{PA-G+} & 227.1 & \textbf{4.2} & 39.2 & 66.4 & 55.3 & 42.9 & 45.6 & 10.0 \\
 & \textsf{CAR+} & \textbf{62.0} & \textbf{4.2} & \textbf{9.0} & \textbf{23.1} & \textbf{19.9} & 18.5 & \textbf{18.3} & \textbf{6.0} \\
\midrule
\multicolumn{10}{@{}l}{$\rho = 10\,\%$} \\
\cmidrule(r){1-2}
\multirow{4}{*}{$\underline{z}_f$} & \textsf{PA-G} & 57.66 & 264.21 & 35.17 & 117.35 & 166.57 & 233.66 & 60.64 & -- \\
 & \textsf{CAR} & 57.67 & 260.40 & 35.17 & 117.36 & 166.58 & 233.68 & 60.65 & -- \\
 & \textsf{PA-G+} & 57.66 & 264.21 & 35.17 & 117.35 & 166.57 & 233.66 & 60.64 & 951.82 \\
 & \textsf{CAR+} & 57.67 & 264.22 & 35.17 & 117.36 & 166.58 & 233.68 & 60.65 & -- \\
\addlinespace[2pt]
\multirow{4}{*}{$\overline{z}_f$} & \textsf{PA-G} & 3606.1 & 9534.6 & 571.0 & 6446.8 & 10516.6 & 10227.6 & 3907.6 & 50069.5 \\
 & \textsf{CAR} & 101.3 & 286.1 & 40.6 & 146.6 & 207.7 & 286.1 & 72.3 & 1024.0 \\
 & \textsf{PA-G+} & 886.7 & 395.3 & 88.5 & 916.3 & 974.2 & 697.1 & 425.5 & 7424.0 \\
 & \textsf{CAR+} & 98.9 & 286.1 & 40.4 & 146.6 & 207.7 & 285.0 & 72.3 & 1024.0 \\
\addlinespace[2pt]
 & $\bar U$ & 156.0 & 286.3 & 48.5 & 178.1 & 244.7 & 314.0 & 81.1 & 1052.2 \\
\addlinespace[2pt]
\multirow{4}{*}{gap (\,\%)} & \textsf{PA-G} & 170.4 & \textbf{8.3} & 38.0 & 51.8 & 46.9 & 34.4 & 33.7 & 10.5 \\
 & \textsf{CAR} & 75.7 & \textbf{8.3} & 15.3 & \textbf{24.9} & \textbf{24.7} & 22.4 & \textbf{19.2} & \textbf{7.6} \\
 & \textsf{PA-G+} & 170.4 & \textbf{8.3} & 38.0 & 51.8 & 46.9 & 34.4 & 33.7 & 10.5 \\
 & \textsf{CAR+} & \textbf{71.5} & \textbf{8.3} & \textbf{14.8} & \textbf{24.9} & \textbf{24.7} & \textbf{22.0} & \textbf{19.2} & \textbf{7.6} \\
\midrule
\multicolumn{10}{@{}l}{$\rho = 15\,\%$} \\
\cmidrule(r){1-2}
\multirow{4}{*}{$\underline{z}_f$} & \textsf{PA-G} & 69.63 & 266.07 & 37.04 & 128.34 & 178.98 & 253.10 & 64.65 & -- \\
 & \textsf{CAR} & 65.42 & 265.68 & 37.05 & 128.35 & -- & 253.12 & 64.65 & -- \\
 & \textsf{PA-G+} & 69.63 & 266.07 & 37.04 & 128.34 & 178.98 & 253.10 & 64.65 & 977.53 \\
 & \textsf{CAR+} & 69.49 & 265.98 & 37.05 & -- & -- & 252.29 & 64.65 & -- \\
\addlinespace[2pt]
\multirow{4}{*}{$\overline{z}_f$} & \textsf{PA-G} & 3722.0 & 10055.4 & 627.1 & 7028.3 & 10521.7 & 10554.1 & 4074.1 & 51194.8 \\
 & \textsf{CAR} & 119.0 & 298.6 & 45.7 & 162.2 & 230.1 & 316.6 & 78.2 & 1072.9 \\
 & \textsf{PA-G+} & 937.4 & 456.8 & 108.5 & 972.4 & 1063.7 & 753.5 & 473.9 & 8239.4 \\
 & \textsf{CAR+} & 124.2 & 298.5 & 45.7 & 162.2 & 230.1 & 316.7 & 78.2 & 1072.9 \\
\addlinespace[2pt]
 & $\bar U$ & 163.6 & 298.7 & 51.3 & 186.5 & 255.8 & 328.6 & 84.6 & 1100.0 \\
\addlinespace[2pt]
\multirow{4}{*}{gap (\,\%)} & \textsf{PA-G} & 134.9 & 12.3 & 38.5 & 45.3 & 42.9 & 29.8 & 30.8 & 12.5 \\
 & \textsf{CAR} & \textbf{70.8} & \textbf{12.2} & \textbf{23.4} & \textbf{26.3} & \textbf{28.6} & \textbf{25.1} & \textbf{20.9} & \textbf{9.8} \\
 & \textsf{PA-G+} & 134.9 & 12.3 & 38.5 & 45.3 & 42.9 & 29.8 & 30.8 & 12.5 \\
 & \textsf{CAR+} & 78.3 & \textbf{12.2} & \textbf{23.4} & \textbf{26.3} & \textbf{28.6} & \textbf{25.1} & \textbf{20.9} & \textbf{9.8} \\
\midrule
\multicolumn{10}{@{}l}{$\rho = 20\,\%$} \\
\cmidrule(r){1-2}
\multirow{4}{*}{$\underline{z}_f$} & \textsf{PA-G} & 78.77 & 267.85 & 38.34 & 136.93 & 189.30 & 268.10 & 67.49 & -- \\
 & \textsf{CAR} & 78.78 & 261.53 & 38.34 & 136.94 & -- & 268.12 & 67.50 & -- \\
 & \textsf{PA-G+} & 78.77 & 267.85 & 38.34 & 136.93 & 189.30 & 268.10 & 67.49 & 998.70 \\
 & \textsf{CAR+} & 78.78 & -- & 38.34 & 135.25 & -- & 260.29 & 67.50 & -- \\
\addlinespace[2pt]
\multirow{4}{*}{$\overline{z}_f$} & \textsf{PA-G} & 3814.9 & 10424.3 & 613.8 & 7235.6 & 11051.3 & 10867.7 & 4152.8 & 52295.9 \\
 & \textsf{CAR} & 134.1 & 311.8 & 49.0 & 174.2 & 249.1 & 337.1 & 83.3 & 1141.2 \\
 & \textsf{PA-G+} & 981.4 & 517.3 & 119.8 & 1038.8 & 1156.5 & 811.6 & 499.4 & 8848.3 \\
 & \textsf{CAR+} & 134.1 & 311.8 & 49.0 & 174.2 & 249.1 & 337.1 & 83.3 & 1122.1 \\
\addlinespace[2pt]
 & $\bar U$ & 170.5 & 311.9 & 53.4 & 194.8 & 266.9 & 342.4 & 88.0 & 1147.9 \\
\addlinespace[2pt]
\multirow{4}{*}{gap (\,\%)} & \textsf{PA-G} & 116.4 & 16.5 & 39.2 & 42.2 & 41.0 & 27.7 & 30.4 & 14.9 \\
 & \textsf{CAR} & \textbf{70.3} & \textbf{16.4} & \textbf{27.8} & \textbf{27.2} & \textbf{31.6} & \textbf{25.7} & \textbf{23.5} & 14.3 \\
 & \textsf{PA-G+} & 116.4 & 16.5 & 39.2 & 42.2 & 41.0 & 27.7 & 30.4 & 14.9 \\
 & \textsf{CAR+} & \textbf{70.3} & \textbf{16.4} & \textbf{27.8} & \textbf{27.2} & \textbf{31.6} & \textbf{25.7} & \textbf{23.5} & \textbf{12.4} \\
\bottomrule
\end{tabular}
\end{adjustbox}
\end{table}

On the primal side, \textsf{PA-G+} finds a feasible solution in all
32~instances, \textsf{PA-G} in~29, \textsf{CAR} in~26, and \textsf{CAR+} in~24; the
primal bounds of \textsf{CAR} and \textsf{CAR+} are up to $6\,\%$ below
$z_{\mathrm{best}}$. The methods therefore differ mainly in their dual
bounds.
Within the time limit, the dual bounds of \textsf{PA-G} and \textsf{PA-G+} never
improve on~$\bar U$, so both attain the gaps given by the analytic
bound, between $4.2\,\%$ and $227\,\%$. \textsf{CAR} and \textsf{CAR+} improve on~$\bar U$
in all 32~instances; their gaps range from $4.2\,\%$ to $78.3\,\%$. \textsf{CAR+} and
\textsf{CAR} attain the same displayed gap in 24~instances, and the largest
difference is $7.5$~percentage points, in favour of \textsf{CAR}
(\codename{biopsy}, $\rho = 15\,\%$). Compared with \textsf{PA-G+}, \textsf{CAR+} reduces
the gap by a factor of up to $4.35$ (median~$1.6$); the improvement
decreases with~$\rho$ and is negligible on \codename{Smarket}.

The additional bounds lower the dual bounds of \textsf{PA-G+} by about one
order of magnitude, although not below~$\bar U$, whereas those of \textsf{CAR+}
barely change. This difference reflects the structure of the two
formulations. In~\eqref{PA_2}, the training points enter through the
sigmoid and the softplus, whose relaxations improve with the much
narrower bounds on~$\theta^{tr}$ (Table~\ref{tab:bound_widths}).
In~\eqref{PA_3}, they enter only through affine constraints and the
convex functions~$\phi$ and~$\phi^*$, which are relaxed well regardless
of the bounds; the only nonconvexity, \eqref{eq:P3_bilinear}, involves
the poisoning points, whose bounds are hardly tightened.
\begin{figure}[t]
  \centering
  \includegraphics[width=\textwidth]{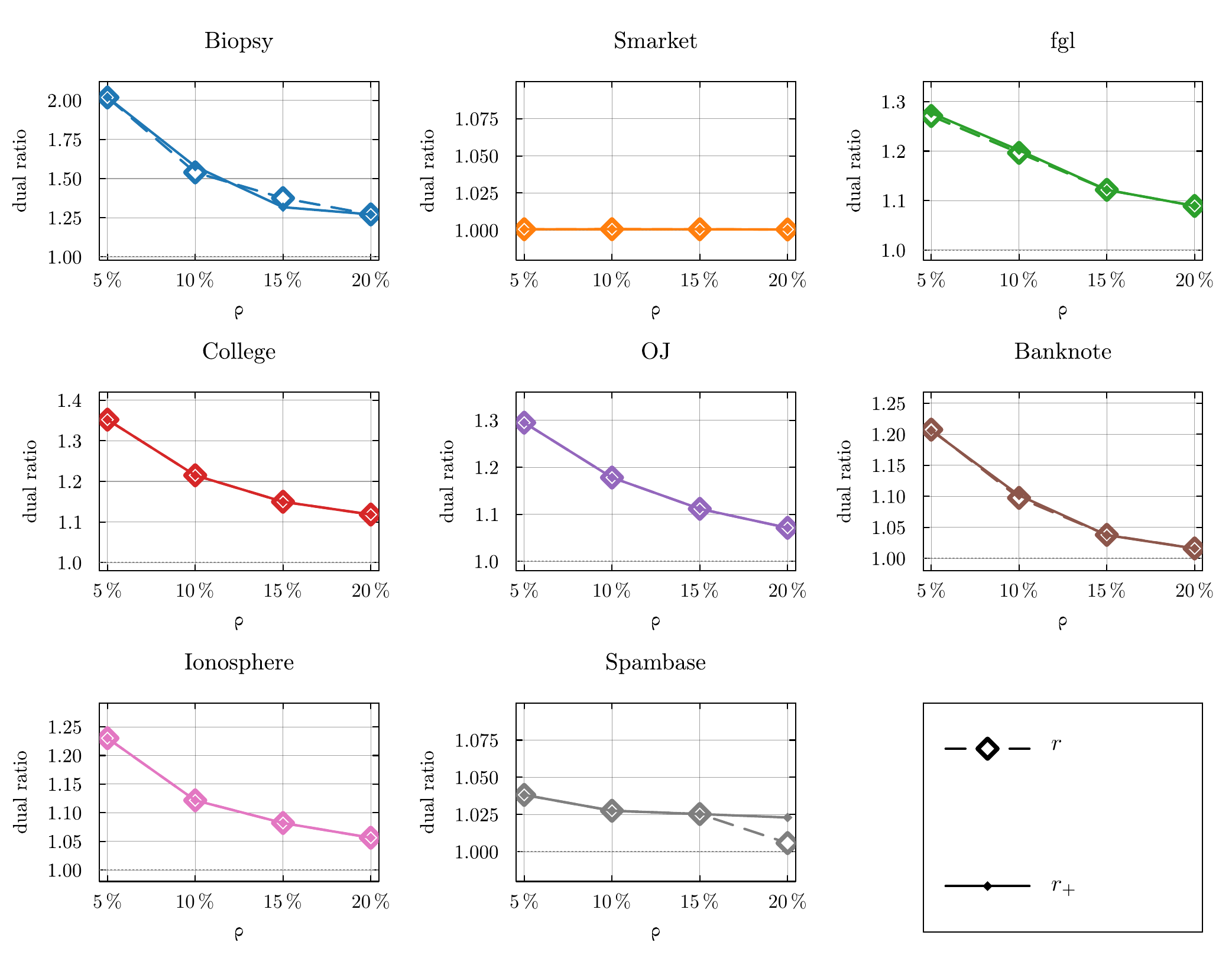}
  \caption{Dual-bound ratios~\eqref{eq:ratio} for each budget~$\rho$:
    $r$ (hollow markers) and $r_+$ (filled markers) as in \eqref{eq:ratio}.
    Note the different scales of the vertical axes.}
  \label{fig:ratio_grid}
\end{figure}

Figure~\ref{fig:ratio_grid} shows the dual-bound ratios~\eqref{eq:ratio}.
Both are at least~$1$ in every instance, so \textsf{CAR+} never yields a weaker
bound than \textsf{PA-G+}. They lie between~1.00 and~2.02 and decrease
with~$\rho$ in every dataset except \codename{Smarket}, where they
remain at~1.00; on \codename{Spambase}, they are at most~1.04. $r$
and~$r_+$ differ by at most~0.05.


\section{Conclusion}
\label{sec:conclusion}

We have developed a convex-analytic reformulation for poisoning attacks
against $\ell_2$-regularized logistic regression. By exploiting convex
conjugacy and a change of variables, we obtain a concave maximization
objective and affine stationarity conditions, with nonconvexity
isolated in bounded bilinear constraints. We derive valid variable
bounds and convex preprocessing problems for tightening them.

Our experiments on eight binary classification datasets show that
attacks obtained by local optimization degrade predictive performance on most
datasets. The convex-analytic formulation substantially improves the
dual bounds reported by \textsf{SCIP} compared with the direct stationarity
formulation, while the reported primal bounds are similar.
Nevertheless, no run certifies global optimality within the time
limit of $3600$\,s.

Future research includes deriving tighter bounds on the poisoning
variables and strengthening the relaxations of the remaining bilinear
terms to reduce the optimality gaps. The reformulation relies only on
the convex conjugate of the loss and on the stationarity conditions of
the lower-level problem, so it may extend to other classifiers trained
by regularized convex loss minimization, such as multinomial logistic
regression or other generalized linear models. Finally, global bounds
on the damage an attacker can cause are a natural starting point for
defenses against poisoning. They can certify the worst-case effect of a
given budget and guide the design and evaluation of robust training
methods.


\section*{Use of AI}

OpenAI's Codex was used to assist with mathematical and editorial review,
consistency checks between the manuscript and the accompanying code,
and the drafting of text. Anthropic's Claude Code was used to help with the
implementation of the models.
The authors retain full responsibility for
verifying all AI-assisted content and for the final manuscript.


\printbibliography

\appendix

\end{document}